\documentclass[pdflatex,sn-mathphys-num]{sn-jnl}

\usepackage{graphicx}%
\usepackage{multirow}%
\usepackage{amsmath,amssymb,amsfonts}%
\usepackage{amsthm}%
\usepackage{mathrsfs}%
\usepackage[title]{appendix}%
\usepackage{xcolor}%
\usepackage{textcomp}%
\usepackage{manyfoot}%
\usepackage{booktabs}%
\usepackage{float}%
\usepackage{hyperref}%

\begin{document}

\title[Diagnosing Verbalized VLM Confidence for Selective Prediction]{Diagnosing Verbalized Confidence in a Medical Vision-Language Model for Selective Prediction on Colorectal Histopathology Patches}

\author[1,2]{\fnm{Anay} \sur{Sinhal}}\email{sinhal.anay@ufl.edu}

\author[3]{\fnm{Amit} \sur{Sinhal}}\email{amit.sinhal@jklu.edu.in}

\author[4]{\fnm{Ashish} \sur{Pandey}} \email{ashish.pandey@jaipur.manipal.edu}

\author*[5]{\fnm{Arpana} \sur{Sinhal}}\email{arpana.sinhal@jaipur.manipal.edu}

\affil[1]{\orgdiv{Department of Computer and Information Science and Engineering}, \orgname{University of Florida}, \orgaddress{\city{Gainesville, FL}, \country{U.S.A}}}

\affil[2]{\orgdiv{Department of Medicine}, \orgname{University of Florida}, \orgaddress{\city{Gainesville, FL}, \country{U.S.A}}}

\affil[3]{\orgdiv{Department of Computer Science and Engineering}, \orgname{JK Lakshmipat University}, \orgaddress{\city{Jaipur}, \country{India}}}

\affil[4]{\orgdiv{Department of Data Science and Engineering}, \orgname{Manipal University Jaipur}, \orgaddress{\city{Jaipur}, \country{India}}}

\affil*[5]{\orgdiv{Department of Computer Applications}, \orgname{Manipal University Jaipur}, \orgaddress{\city{Jaipur}, \country{India}}}

\abstract{Flagging uncertain histopathology patches for specialist review needs a confidence that is higher when the model is right. We test whether an open medical vision-language model emits one. MedGemma-27b-it and its base Gemma-3-27b-it were run zero-shot under one prompt on 10,004 PathMNIST validation patches and 7,180 patches of a patient-disjoint external cohort. Both are too inaccurate to deploy (40.6\% and 35.5\%); the object of study is the confidence. MedGemma's verbalized confidence puts 96\% of patches at one value; its confidence-correctness AUROC is 0.499 and 0.528, and routing by it moves selective-accuracy AUC by $-0.004$ and $+0.037$. Gemma-3 spreads its confidence over nine values, reaches AUROC 0.70, and gains $+0.114$ and $+0.115$. MedGemma's label-token probabilities do not rescue it: their validation advantage reverses where the top-20 alternatives include every class. On 1,800-patch subsets of each cohort, stratified on correctness and reweighted to the full splits, agreement across five prompt variants beats random by $+0.123$ and $+0.106$, and wording-only paraphrases give $+0.100$ and, externally and in direction only, $+0.054$; agreement's external margins over confidence do not survive slide-level clustering. Under one restructured prompt variant, MedGemma's stated confidence routes ($+0.112$, $+0.141$), so the failure is tied to a prompt chosen for macro F1.}

\keywords{histopathology, vision-language model, calibration, selective prediction, uncertainty-aware routing, colorectal, verbalized confidence, cross-cohort generalization}

\maketitle

\section{Introduction}\label{sec:intro}

Colorectal cancer is among the most prevalent malignancies worldwide, and tissue diagnosis depends substantially on histopathological review of H\&E-stained biopsies. Interpreting ambiguous cases benefits from subspecialist consultation, but gastrointestinal pathologists are unevenly distributed. Community hospitals in underserved regions often rely on a single generalist pathologist, creating a bottleneck for complex cases.

\begin{figure}[H]
\centering
\includegraphics[width=\textwidth]{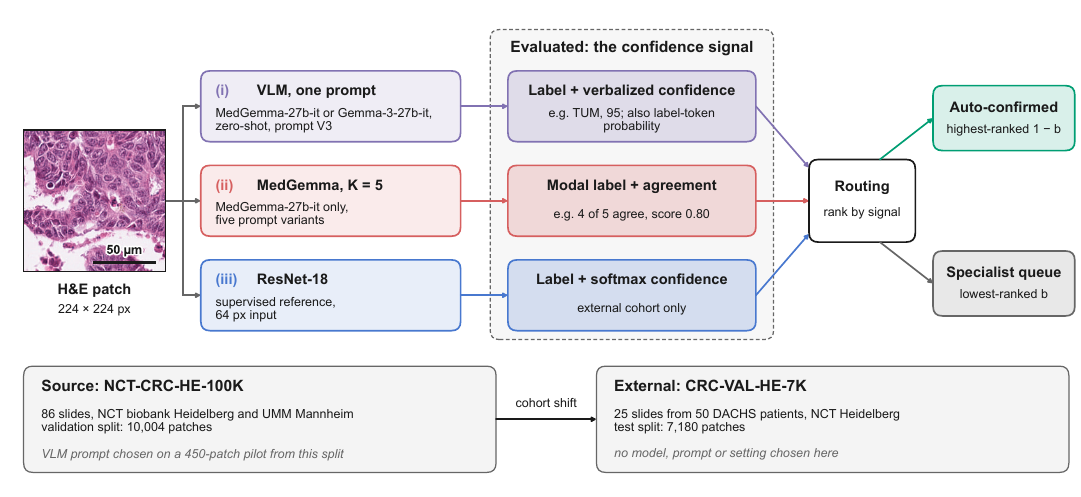}
\caption{The selective-prediction setting studied here. A patch from the source cohort (NCT-CRC-HE-100K) or the patient-disjoint external cohort (CRC-VAL-HE-7K) is classified by MedGemma-27b-it or Gemma-3-27b-it under one prompt (i), by MedGemma-27b-it alone under five prompt variants (ii), or by a supervised ResNet-18 (iii), and each route yields a label with a confidence signal. The routing policy ranks patches by that signal, sends the lowest-ranked fraction $b$ to a simulated specialist queue and auto-confirms the rest. The object of evaluation is the confidence signal (dashed box). The patch shown is a real validation patch from NCT-CRC-HE-100K \cite{kather2018nct} (CC BY 4.0), PathMNIST validation index 6404 (scale bar 50~$\mu$m; the source release gives 0.5~$\mu$m per pixel), and both example outputs are what MedGemma-27b-it returned for it: TUM at confidence 95 under V3, and TUM from four of the five variants.}\label{fig:pipeline}
\end{figure}

A second-read triage layer is one way to address this gap (Figure~\ref{fig:pipeline}). Rather than replacing the pathologist, such a layer would auto-confirm routine patches with high confidence while routing the genuinely uncertain ones to a specialist queue. This is a selective-prediction problem \cite{elyaniv2010foundations,geifman2017selective}, and it depends on one property above all: a confidence signal that is higher on the cases the model gets right than on the cases it gets wrong. Whether an open medical VLM emits such a signal is an empirical question, and it is the question this paper is about. We do not propose a deployable triage system; at the accuracy these zero-shot models reach, none is possible. We instead diagnose the confidence.

MedGemma-27b-it, released by Sellergren et al. as an open VLM fine-tuned on curated biomedical data \cite{sellergren2025medgemma}, operates at the patch level and produces a tissue label alongside a verbalized confidence value, an integer it writes in its own response. Its general-purpose base, Gemma-3-27b-it \cite{gemmateam2025gemma3}, produces the same output under an identical prompt, so the two form a matched pair: one architecture, one prompt, one endpoint. The pair shows how the tuned model's confidence signal differs from its base's. It does not attribute that difference to medical tuning alone, because the endpoint does not expose which checkpoint it serves (Section~\ref{subsec:consistency-methods}).

Two things were unsettled when we started. Verbalized VLM confidence on histopathology classification had been characterized on small balanced subsets rather than on a full evaluation split, and never with a direct measurement of how much it says about correctness. And no medically tuned model had been compared with its own base on the confidence signal itself, on one task under one prompt.

We evaluate both models on all 10,004 validation and 7,180 external patches from PathMNIST, with a supervised ResNet-18 as an external-cohort reference, and report a dissociation. MedGemma's confidence carries no usable information about correctness and does not route; Gemma-3's carries a small signal that routes consistently on both cohorts. Where the stated number fails, agreement across prompt variants routes, and so does agreement across paraphrases that change only the wording. On the same patches, one restructured prompt draws from MedGemma a stated confidence that routes about as well as Gemma-3's, though its AUROC is lower. The mechanism is degeneracy rather than misdirected confidence (Section~\ref{subsec:disc-routing}).

\section{Related Work}\label{sec:related}

\textbf{Colorectal H\&E datasets.}
Kather et al. assembled NCT-CRC-HE-100K, 100,000 non-overlapping 224-pixel H\&E patches from 86 colorectal tissue slides labeled across nine tissue types \cite{kather2018nct}. The companion dataset CRC-VAL-HE-7K, a patient-disjoint collection cut from a separate set of slides, was released alongside a survival prediction study \cite{kather2019predicting} and is the external cohort here (Section~\ref{subsec:dataset} states its provenance and the sense in which the two collections differ). Yang et al. packaged both as PathMNIST within MedMNIST v2 \cite{yang2023medmnist}, standardizing the train/val/test splits used throughout this work; its predecessor \cite{yang2021medmnist} established the same lightweight evaluation philosophy across a broader set of medical imaging tasks.

\textbf{VLMs in computational pathology.}
Chen et al. trained UNI on over 100 million patches from more than 100,000 whole-slide images, achieving strong performance across classification and survival tasks via self-supervised pretraining followed by task-specific fine-tuning \cite{chen2024uni}. That approach requires at least some labeled data at deployment. MedGemma \cite{sellergren2025medgemma} and its general-purpose base Gemma 3 \cite{gemmateam2025gemma3} classify via text generation without local fine-tuning, which is the property that makes them relevant for low-resource deployment. A related line of work has coupled pathology images with naturally occurring text supervision rather than task labels: Huang et al. trained PLIP on medical-Twitter image-text pairs to obtain a zero-shot pathology vision-language model \cite{huang2023plip}, and broader surveys of foundation models for generalist medical AI make the same low-resource argument for text-generation-based clinical tools \cite{moor2023foundation}. Med-PaLM demonstrated that general-purpose clinical language models can encode substantial medical knowledge while still falling short of specialist-level reliability \cite{singhal2023medpalm}. Whether the confidence attached to a zero-shot VLM prediction is reliable enough to drive automated triage is the question we address.

\textbf{Calibration.}
Guo et al. showed that well-trained deep classifiers are frequently overconfident and that temperature scaling reliably reduces expected calibration error \cite{guo2017calibration}. Their method targets full softmax output. Applying it to a VLM that emits only a scalar verbalized confidence requires a log-odds reformulation (Section~\ref{subsec:calibration-methods}). Nixon et al. later showed that ECE itself is sensitive to binning choices and proposed complementary calibration diagnostics for a more complete picture of miscalibration \cite{nixon2019measuring}.

\textbf{Uncertainty in medical VLMs.}
Concurrent with this work, Sadanandan and Behzadan studied MedGemma-4B-IT and LLaVA-RAD-7B on binary presence questions over chest radiographs, showing that single-pass predictive entropy predicts which yes/no answers flip when a clinically equivalent question is rephrased, at AUROC above 0.82 on both MIMIC-CXR and PadChest \cite{sadanandan2026entropy}. Their setting differs from ours on every axis that bears on the conclusion: radiographs against colorectal H\&E patches, binary presence against nine-way tissue classification, locally hosted weights with access to the answer-head logit margin against a black-box remote endpoint, and a logit-derived signal against the integer the model writes. The two results complement each other. Where logits are available the token distribution already carries the instability signal; we show that when they are not, and the verbalized number is degenerate, the instability is still measurable from the responses alone.

Two further studies bracket the question. Khanmohammadi et al. benchmarked nine confidence estimators, among them prompt-based self-reports and a prompt ensemble, across five open-weight VLMs including Gemma-3-27B on three medical VQA sets \cite{khanmohammadi2026triage}. At a 20\% error tolerance their best estimator could safely defer a quarter of radiology cases and next to none of pathology, and they conclude that a good estimator lets a competent model defer well but cannot manufacture reliability the model lacks. We reach the same wall from a different direction. What their design cannot address is whether medical fine-tuning itself changes the confidence signal, since their five models are not a tuned-and-base pair and their estimators are probe-derived or forced-choice rather than the free-form number a model volunteers. Senoglu et al. take the constructive route, fine-tuning MedGemma-4B-IT with a calibration-aware objective and cutting calibration error substantially on medical VQA \cite{senoglu2026calibration}; that this helps is consistent with our reading of the deficit as a property of the released model under a fixed prompt rather than something that resists correction.

One published result cuts against agreement as a signal. Sadanandan and Behzadan show that a medical VLM can be perfectly consistent under paraphrase while barely using the image, answering from language priors instead, with 98.5\% of samples consistent but not image-reliant at a 1.5\% flip rate \cite{sadanandan2026dangerous}. Agreement across paraphrases is therefore not automatically evidence of anything, and Section~\ref{subsec:consistency-results} reports the image controls that test it on our data.

\textbf{Selective prediction.}
El-Yaniv and Wiener set out the risk-coverage view of selective classification, in which a classifier abstains on uncertain samples to reach a target accuracy on the covered subset \cite{elyaniv2010foundations}, and Geifman and El-Yaniv applied it to deep networks \cite{geifman2017selective}. We adopt that framework directly and evaluate accuracy on the auto-confirmed subset across a range of routing budgets. Their later SelectiveNet trains the reject option jointly with the classifier \cite{geifman2019selectivenet}, which is not available to us: we query a fixed remote model and cannot modify its training objective, so the abstention rule has to be built from whatever signal the model already emits.

\textbf{Domain shift.}
Tellez et al. quantified how strongly stain color variation alone degrades convolutional network accuracy in computational pathology, and found that augmentation and normalization only partially compensate \cite{tellez2019quantifying}. Stacke et al. proposed explicit metrics for the magnitude of domain shift between histopathology datasets, making the size of the gap measurable rather than assumed \cite{stacke2021measuring}. Campanella et al. worked at a different scale, training weakly supervised whole-slide classifiers on more than 44,000 slides and reporting clinical-grade accuracy without pixel-level annotation \cite{campanella2019clinical}. All three studies concern supervised visual models; Section~\ref{subsec:crosscenter} measures how far a zero-shot VLM's accuracy moves under a cohort shift that survives stain normalization.

\textbf{Consistency and self-verification.}
A separate line of work has used repeated or varied querying of a language model as a signal distinct from its stated confidence. Wang et al. showed that sampling multiple reasoning paths and taking the majority answer (self-consistency) improves chain-of-thought accuracy without any changes to the model itself \cite{wang2023selfconsistency}. Manakul et al. proposed SelfCheckGPT, which detects hallucinated content by checking whether repeated samples from the same model agree with each other, entirely without access to external knowledge or model internals \cite{manakul2023selfcheckgpt}. This idea has begun to reach medical vision-language models: Wang et al. measured semantic consistency across sampled radiology reports and used it to reject the least factual generations, again without touching model internals \cite{wang2024semantic}. We apply the same principle to patch classification, measuring agreement across prompt variants in the setting where the model's stated confidence carries no routing signal. Kadavath et al. found that language models' verbalized confidence is often poorly calibrated even though internal probability estimates can be well calibrated \cite{kadavath2022know}, and Xiong et al. and Tian et al. both reported that prompting strategies for eliciting calibrated confidence from instruction-tuned language models are unreliable across tasks \cite{xiong2024uncertainty,tian2023askcalibration}.

\section{Methods}\label{sec:methods}

\subsection{Dataset}\label{subsec:dataset}

PathMNIST provides the experimental substrate. The nine tissue classes are adipose (ADI), background (BACK), debris (DEB), lymphocytes (LYM), mucus (MUC), smooth muscle (MUS), normal colon mucosa (NORM), cancer-associated stroma (STR), and colorectal adenocarcinoma epithelium (TUM). MedMNIST v2 \cite{yang2023medmnist} defines the splits but distributes images at $28 \times 28$; the 64~px and 224~px releases we use come from the later MedMNIST+ collection \cite{doerrich2025medmnistplus}. Both VLMs receive $224 \times 224$ RGB patches and the CNN reference is trained at 64~px. Comparing the two releases' label arrays element-wise across all three splits confirms that they index the same patches in the same order, so the VLM and CNN evaluations cover identical patches.

\textbf{Provenance, and what kind of shift this is.} The training split (89,996 patches) and validation split (10,004 patches) come from NCT-CRC-HE-100K \cite{kather2018nct}, which Kather et al. assembled from 86 H\&E slides drawn from two institutions: the NCT biobank in Heidelberg and the pathology archive of the University Medical Center Mannheim. The external test split (7,180 patches from CRC-VAL-HE-7K \cite{kather2019predicting}) was cut from 25 H\&E slides of tissue from 50 colorectal adenocarcinoma patients \cite{kather2018nct} in the DACHS study, banked at NCT Heidelberg and assembled so that no patient overlaps the first collection. This is not a two-hospital separation. The source is not one center, and the target draws on an institution that also contributes to the source. What the collections do differ in is cohort, collection period, and staining and scanning batch, with disjoint patients, so we describe the shift as a cohort-and-batch shift throughout. Both were also colour-normalized with the Macenko method before release \cite{kather2018nct,kather2019predicting}, which bounds it further: what remains is whatever survives a stain normalizer applied separately to each collection, plus the scanner and cohort differences normalization does not touch. Robustness to staining is therefore tested here only at that reduced magnitude.

Cluster counts follow from the same provenance. Validation patches nest within 86 slides; external-test patches nest within 25 slides and 50 patients. The two units are not interchangeable, and we use the slide, because staining, scanner and batch attach to it and because 25 clusters is the more conservative choice. Section~\ref{subsec:routing-methods} states what that does to every interval we quote.

\textbf{Class distribution.} The two splits are unbalanced and unbalanced differently, which matters for any chance-level comparison (Table~\ref{tab:classdist}). The majority class is TUM at 14.3\% on validation and ADI at 18.6\% on the external test set, both above the 11.1\% uniform rate.

\begin{table}[h]
\caption{Per-class patch counts and shares for the validation and external test splits. The majority-class rate, not the 11.1\% uniform rate, is the relevant reference for an above-chance claim, and it differs between the two splits.}\label{tab:classdist}
\begin{tabular}{@{}lcccc@{}}
\toprule
 & \multicolumn{2}{c}{Validation (NCT-CRC-HE-100K)} & \multicolumn{2}{c}{External test (CRC-VAL-HE-7K)} \\
\cmidrule(lr){2-3}\cmidrule(lr){4-5}
Class & $n$ & \% & $n$ & \% \\
\midrule
ADI & 1,041 & 10.4 & 1,338 & 18.6 \\
BACK & 1,057 & 10.6 & 847 & 11.8 \\
DEB & 1,152 & 11.5 & 339 & 4.7 \\
LYM & 1,156 & 11.6 & 634 & 8.8 \\
MUC & 890 & 8.9 & 1,035 & 14.4 \\
MUS & 1,354 & 13.5 & 592 & 8.2 \\
NORM & 877 & 8.8 & 741 & 10.3 \\
STR & 1,045 & 10.4 & 421 & 5.9 \\
TUM & 1,432 & 14.3 & 1,233 & 17.2 \\
\midrule
Total & 10,004 & 100 & 7,180 & 100 \\
\botrule
\end{tabular}
\end{table}

\textbf{Use of the external test split.} The external split is not a single-touch held-out evaluation. It enters five analyses: one full-scale inference pass per model; calibration transfer, applying the validation-fitted temperature without refitting; the routing analysis; the cross-cohort comparison; and the K=5 consistency transfer on an 1,800-patch subset. The last of these selects patches using test-set outcomes, since the subset is stratified on whether the full-scale test prediction was correct, so it is a follow-up analysis rather than a held-out evaluation and we do not present it as one. No model, prompt, or hyperparameter was selected on the external split.

All five models are evaluated on all 10,004 validation patches and all 7,180 test patches. Prompt selection used a pilot of 50 validation patches per class, the same 450 for every prompt version, and no external patch. The pilot patches were not withheld from the later validation analyses: 370 of them fall in MedGemma's calibration set and 80 in its held-out evaluation partition (367 and 83 for Gemma-3), and 78 in the validation consistency subset. Every validation headline was recomputed with the 450 removed, and none changed sign or whether its interval excludes zero (Supplementary Information). Full-scale accuracy exceeds the balanced-pilot estimate, and class frequency does not explain the difference. BACK is 10.6\% of the validation split, below the uniform share, and the two largest classes are TUM and MUS. MedGemma's best per-class F1 scores are on BACK and LYM (Table~\ref{tab:perclass}), its easiest classes but not its most frequent ones.

\subsection{Models}\label{subsec:models}

\textbf{MedGemma-27b-it} is a 27B-parameter multimodal model fine-tuned from Gemma 3 on curated biomedical data \cite{sellergren2025medgemma}. We query it via an OpenAI-compatible API at temperature 0.0, with the V3 prompt that lists all nine tissue classes with brief histological cues (for example, ``smooth muscle: elongated eosinophilic fibers, cigar-shaped nuclei'') and instructs the model to respond with a label, integer confidence (0--100), and one sentence of reasoning. The label is read from the response's LABEL line and the confidence from its CONFIDENCE line. A response with several LABEL lines names more than one class and is scored as unknown, as is a label outside the class list; an unknown answer counts as an error. All 17,184 of MedGemma's full-scale responses name one listed class.

Three groups of runs stored only the parsed label and confidence, not the response text: the K=5 runs of the original variant family on both splits, the V5 family, and the label-token runs (Sections~\ref{subsec:consistency-methods} and~\ref{subsec:logprob-methods}). Their labels come from an earlier parser that searched the whole response, reasoning sentence included, for the nine class names in a fixed priority order and returned the first that appeared anywhere, so that ``background \dots{} no tissue or debris'' was read as debris. On the runs that kept their text the two parsers disagree on 193 of MedGemma's 10,004 validation labels and 123 of its 7,180 external labels (1.9\% and 1.7\%), almost all of them background read as debris, and on 54 and 13 of Gemma-3's. Every result reported here for a run that kept its text uses the LABEL line. The runs without text keep the earlier labels, and a comparison between one of them and a run that kept its text uses the earlier rule on both sides, so that the change of parser is not counted as a difference between the runs.

The prompt registry defines six versions (Table~\ref{tab:prompts}). Five were evaluated in 450-sample pilots. V1 was drafted and never run, so it carries no macro F1 and the comparison effectively begins at V2. V3 achieved the best macro F1 (0.301) and is used for all full-scale evaluations. The ablation showed that disambiguation rules targeting the three zero-F1 classes (MUC, NORM, STR) raised F1 on mucus and normal mucosa, left stroma at 0.00 under both V4 and V4B, and reduced macro F1 overall, a pattern consistent with the model shifting its over-prediction bias across classes rather than resolving it.

V3 was chosen on macro F1 alone. Confidence quality was never a selection criterion and was not inspected until the prompt had been fixed.

\begin{table}[h]
\caption{Prompt versions in the registry, with pilot macro F1 on the 450-sample balanced subset. V3 was selected on macro F1 and used for every full-scale run.}\label{tab:prompts}
\begin{tabular}{@{}llc@{}}
\toprule
Version & Description & Pilot macro F1 \\
\midrule
V1 & Minimal class list, no histological cues & not evaluated \\
V2 & Class list with brief cues & 0.213 \\
V3 & V2 plus per-class diagnostic cues (selected) & 0.301 \\
V4 & V3 plus disambiguation rules for MUC, NORM, STR & 0.283 \\
V4B & V4 with stricter tie-breaking instructions & 0.227 \\
V5 & Restructured instruction set with step-by-step rules & 0.294 \\
\botrule
\end{tabular}
\end{table}

\textbf{Gemma-3-27b-it} is the general multimodal base from which MedGemma was fine-tuned \cite{gemmateam2025gemma3}. We use identical prompts and inference settings. The two models then differ in medical tuning and in anything else that separates the two served checkpoints, which the endpoint does not identify, so a difference between them is consistent with an effect of tuning without isolating one. Gemma-3 answered outside the class list on 2 validation and 6 test patches (0.02\% and 0.08\%), naming tissues such as blood vessels or fungal hyphae at stated confidence 0.90 to 0.95, and listed several classes in one response on 16 and 5. All of these count as errors in accuracy, routing and full-split AUROC. The eight off-list answers name no class to be right about and are left out of the calibration split (Section~\ref{subsec:calibration-methods}); the list answers stay in it.

\textbf{ResNet-18} is an 18-layer residual network \cite{he2016resnet} with an ImageNet-pretrained backbone, adapted to nine output classes and trained on the 64~px MedMNIST+ release \cite{doerrich2025medmnistplus} with Adam (learning rate 1e-3, weight decay 1e-4), cosine annealing over 40 epochs, and seed 42.

The CNN has a restricted role. PathMNIST partitions NCT-CRC-HE-100K at the patch level, so patches from every training slide also appear in validation, and the CNN's validation accuracy (0.9971) and every quantity fitted or measured on validation are leakage-inflated. No slide or patient identifier is distributed with the data (Section~\ref{subsec:routing-methods}), so a slide-disjoint split cannot be built. We therefore report the CNN's validation figures for completeness and draw no conclusion from them. The CNN enters the comparisons only through the external cohort, which shares no patient with its training data: there its accuracy, calibration error and routing behaviour are clean measurements of a supervised reference.

\textbf{PLIP} is a CLIP model with a ViT-B/32 image encoder, trained by Huang et al.\ on pathology image-text pairs \cite{huang2023plip}. It classifies by matching an image against class descriptions rather than by generating text, so it separates what the images make visible from what a generation interface reports. We use the public checkpoint (\texttt{vinid/plip}, commit \texttt{67ade53}) in fp16 on the local GPU, in two ways. Zero-shot, each patch is scored against the nine class names under four prompt templates averaged in embedding space, a choice fixed before the run, and the softmax probability of the predicted class is its confidence; a single template is reported beside it in the Supplementary Information. As a probe, a multinomial logistic regression is trained on frozen image embeddings of 19,998 class-stratified training patches, its regularization chosen by five-fold cross-validation on those patches; its validation figures share slides with its training patches, so, as for the CNN, only its external figures are interpreted. Whether PLIP's pretraining images overlap these cohorts is not known. \textbf{QuiltNet}, a second CLIP model with a ViT-B/32 image encoder, trained by Ikezogwo et al.\ on pathology image-text pairs drawn from educational videos \cite{ikezogwo2023quilt}, is run in exactly the same way, with the same templates and probe (\texttt{wisdomik/QuiltNet-B-32}, commit \texttt{8ce7728}); whether its pretraining images overlap these cohorts is not known either.

\subsection{Calibration}\label{subsec:calibration-methods}

We measure Expected Calibration Error with 15 equal-width confidence bins over $[0, 1]$:

\begin{equation}
\text{ECE} = \frac{1}{N} \sum_b n_b \left| \overline{\text{conf}}_b - \overline{\text{acc}}_b \right|
\label{eq:ece}
\end{equation}

\noindent where $n_b$, $\overline{\text{conf}}_b$, and $\overline{\text{acc}}_b$ are the count, mean confidence, and mean accuracy of bin $b$. Because equal-width ECE is sensitive to the binning scheme \cite{nixon2019measuring}, we also report an adaptive ECE with 15 equal-mass bins (equal sample count per bin) and the Brier score $\frac{1}{N}\sum_i (c_i - y_i)^2$, a binning-free proper scoring rule where $c_i$ is the confidence and $y_i \in \{0,1\}$ the correctness of prediction $i$.

For VLMs, only a scalar confidence is available. We apply log-odds temperature scaling: the calibrated confidence is $\sigma(\text{logit}(c) / T)$ where $c$ is the raw confidence and $\sigma$ is the sigmoid function. T is fitted by minimizing binary negative log-likelihood, treating each prediction as a binary correct/incorrect outcome, with a bounded scalar search over $T \in [0.05, 200]$. A stated confidence of 100 is clipped to $1 - 10^{-7}$ before the logit, giving log-odds 16.1; 283 of MedGemma's 8,003 calibration confidences are 100, and none of Gemma-3's. For comparison, two less restricted maps are fitted on the same calibration split: Platt scaling with a slope and an intercept on the log-odds, and isotonic regression. Each is scored on the held-out partition and, unchanged, on the external cohort, with the Brier score split into reliability and resolution over the distinct calibrated values \cite{murphy1973partition}.

Both VLMs reach the upper bound, and on these data that follows from the confidences themselves. Write $z_i = \text{logit}(c_i)$ and $y_i$ for correctness. The derivative of the binary negative log-likelihood with respect to $1/T$ is $\sum_{y_i = 0} z_i\,\sigma(z_i/T) - \sum_{y_i = 1} z_i\,\sigma(-z_i/T)$. When every $z_i$ is positive, the first sum is at least half the log-odds summed over errors and the second at most half the log-odds summed over correct predictions, so the likelihood improves with every increase in T whenever the first total is at least the second. MedGemma meets this condition on its calibration set: its lowest confidence is 0.90, and its log-odds sum to 16,336 over errors against 10,947 over correct predictions. Gemma-3 misses it only because 29 of its 8,001 confidences lie below 0.5. The sums depend on the clip and the condition does not: with clips from $10^{-2}$ to $10^{-7}$, MedGemma's sum over errors stays between 1.45 and 1.49 times its sum over correct predictions. On both calibration sets the binary NLL decreases monotonically over $T \in [0.05, 10^5]$, from 2.091 at $T = 1$ (1.828 with a clip of $10^{-2}$) to 0.695 at $T = 200$ for MedGemma and from 1.494 to 0.694 for Gemma-3, approaching its limit $\ln 2 \approx 0.693$ (Supplementary Information). A fitted T of 200 is therefore a property of the confidences and the search interval rather than a measured temperature, and Section~\ref{subsec:calibration-results} treats it accordingly.

The calibration set is a random 80\% of the validation patches (seed 42): 8,003 of 10,004 for MedGemma and the CNN, and 8,001 of the 10,002 that remain for Gemma-3 once its two off-list answers are set aside. The remaining 2,001 patches form each model's held-out evaluation partition, so the two VLMs' partitions are drawn separately and share 1,637 patches. For cross-cohort transfer we apply the val-fitted T to test samples without re-fitting. For the CNN, T is fitted by minimizing cross-entropy on logit vectors via L-BFGS.

\subsection{Routing Policy}\label{subsec:routing-methods}

At routing budget $b$, the policy routes the $b \cdot N$ lowest-confidence patches to a simulated specialist queue and auto-confirms the remaining $(1-b) \cdot N$ highest-confidence patches. Let $S(b)$ be that auto-confirmed set and $\text{acc}(S(b))$ its accuracy. We summarize the policy by the area under the auto-confirmed-accuracy curve,

\begin{equation}
A_{\text{AUC}} = \int_0^1 \text{acc}(S(b))\, db,
\label{eq:aauc}
\end{equation}

\noindent estimated on the budgets $b_k = k/100$, $k = 0, \dots, 99$. The budget $b = 1$ is dropped because $S(1)$ is empty and its accuracy undefined, and the trapezoid rule is applied to the remaining 100 points with spacing $1/99$, so the estimate is
\begin{equation}
\begin{split}
\hat{A}_{\text{AUC}} &= \frac{1}{0.99} \int_0^{0.99} \text{acc}(S(b))\, db \\
&\approx \frac{1}{99}\left[\tfrac{1}{2}\,\text{acc}(S(b_0)) + \sum_{k=1}^{98} \text{acc}(S(b_k)) + \tfrac{1}{2}\,\text{acc}(S(b_{99}))\right],
\end{split}
\label{eq:aauc-est}
\end{equation}
a trapezoidal average of auto-confirmed accuracy over $[0, 0.99]$ rescaled to unit width. Every $A_{\text{AUC}}$ in this paper is that average. It is bounded by the largest accuracy on the curve, so a near-perfect ranking of a 99.7\%-accurate classifier, such as the CNN on validation, can print as 0.9998 or 1.0000 after rounding. This is the accuracy-oriented complement of the risk-coverage curve \cite{elyaniv2010foundations,geifman2017selective}: we integrate accuracy on the retained set rather than error, so \textbf{higher is better}. A non-informative (random) confidence signal yields $A_{\text{AUC}}$ equal to the overall accuracy $a$ of the set. A perfect ranking, which routes every error before any correct prediction, reaches $(a - a \ln a - 0.01)/0.99$ rather than 1: 0.770 and 0.717 for MedGemma on the two splits and 0.720 and 0.669 for Gemma-3, which is the scale against which their gaps should be read. We refer to $A_{\text{AUC}}$ as the selective-accuracy AUC throughout to avoid confusion with the error-oriented AURC (lower-is-better) that the same framework can also produce.

Three policies are compared: calibrated routing (sorting by calibrated confidence), raw routing (sorting by uncalibrated confidence), and random routing (a uniformly random ordering, whose expected auto-confirmed accuracy is the base accuracy at every budget, so that its $A_{\text{AUC}}$ is that accuracy exactly). Since log-odds temperature scaling is monotone, calibrated and raw VLM routing are identical in ranking. We report $A_{\text{AUC}}$ and the operating point at $b = 0.15$. To test whether a model's selective-accuracy AUC beats random, we compute a paired bootstrap (1{,}000 resamples, patch identity held paired within each resample) of the gap between calibrated and random routing; because random routing AUC equals the base accuracy in expectation, the resampled base accuracy is the paired random reference.

Every gap we print is the difference of two point estimates computed at full precision and rounded once, while the AUCs beside it are rounded to three decimals first, so subtracting printed cells can land 0.001 away; four-decimal values are in the deposited artifacts. Each interval is a paired bootstrap, whose mean differs from the point estimate beside it by at most 0.002 across every contrast we report, and every printed point estimate falls inside its own interval. Gaps are taken against a reference computed on the outcome the signal routes, which for the single-query row is its own correctness rather than the modal vote's.

\textbf{Tie handling.} Verbalized confidence is coarse, so most of the ranking is undetermined: MedGemma's full-scale confidence takes four distinct values on validation and five on the external split, with 96\% of patches at one of them, leaving 92\% of patch pairs tied. Resolving those by row position reports a property of the file rather than of the model. Across 200 random tie orders of the single-query confidence on the 1,800-patch external subset, $A_{\text{AUC}}$ runs from 0.328 to 0.386, a spread of 0.059 that reflects nothing but the ordering. We therefore resolve ties by their expectation over uniformly random tie orders, computed in closed form, and report beside each AUC the tie fraction and the standard deviation tie ordering induces. That standard deviation comes from 200 sampled orders, since the closed form has no spread of its own; the sampled mean lands within 0.0013 of the exact value on every signal, and the exact value is what every table and every bootstrap uses. A regression test in the released code enforces invariance to row permutation.

\textbf{Clustering.} Patches nest within slides (Section~\ref{subsec:dataset}), so the patch-level bootstrap understates uncertainty. A slide-level cluster bootstrap is closed to us: MedMNIST v2 publishes no index from a PathMNIST array position back to its source slide, and the distributed archives hold only images and labels. In its place we report, for every headline interval, how it widens under an assumed intra-cluster correlation $\rho$ through the design effect $1 + (\bar{m} - 1)\rho$, and the break-even $\rho$ at which each conclusion changes (Section~\ref{subsec:routing-results}). With 86 slides on validation and 25 on the external split, mean cluster sizes are 116 and 287 patches on the full splits, 21 and 72 on the 1,800-patch consistency subsets, and 23 and 80 on the label-token samples of Section~\ref{subsec:logprob-methods}. This is sensitivity analysis rather than clustered inference: $A_{\text{AUC}}$ is a functional of the whole ranking rather than a simple mean, so the design effect is an approximation.

\textbf{Multiplicity.} The conclusions rest on 19 contrasts, fixed as one family in the released code: 16 of them after the runs that produced them, two before the V3W and image-control runs, and the last before the external V3W run. They are the verbalized-confidence routing gap of each VLM; agreement against random, against averaged confidence, against single-query confidence routing the modal-vote outcome, and among the three V3 edits alone; and label-token confidence against random and against the verbalized number, each on both splits; plus agreement against random under the V3W family on both splits and on pixel-shuffled patches on validation. Each contrast is tested with a two-sided normal-approximation p value from its bootstrap standard error, and Holm's step-down procedure holds the family-wise error rate at 0.05. Intervals are printed unadjusted, and the adjusted outcome of every contrast is in the Supplementary Information.

\subsection{Consistency-Based Uncertainty}\label{subsec:consistency-methods}

\textbf{Samples and weighting.} The K=5 experiment ran on 1,800 validation patches and, separately, on 1,800 external patches. On each split the draw took up to 100 patches that the full-scale V3 prediction got right and up to 100 that it got wrong from each class (seed 42), filling from the other pool where one fell short, so that correct and incorrect predictions were both represented in every class. That design changes the class mix and the share of correct predictions, so an average over the subset describes the subset rather than the split. Each patch's inclusion probability is known exactly from the size of its cell (class by full-scale correctness), and every cohort-level quantity in Section~\ref{subsec:consistency-results} is a Horvitz-Thompson estimate in which a patch stands for the inverse of that probability. For the selective-accuracy AUC the routing budget becomes a fraction of the weighted total and a tied group straddling the cut is included fractionally; with every weight equal to 1 this is the estimator of Section~\ref{subsec:routing-methods}. Intervals come from a paired bootstrap that resamples inside each selection cell. The weighting recovers what it should: the weighted accuracy of the in-session single query is 0.3422 on the external split, equal to the full-scale accuracy under the labels the subsets were drawn on (Section~\ref{subsec:models}), and 0.4005 on validation against 0.4023, a difference due to the 54 drifted labels described below. Subset values are reported beside the cohort estimates. A family built from the V5 prompt ran earlier, on 29 June 2026, on a different validation sample of 200 patches per class with no stratification on correctness, and is weighted by class share in the same way. Its variants 0, 1, 3 and 4 apply the same edits to V5, and its variant 2 is the same prompt as in the V3 family.

\textbf{The five variants.} For each patch MedGemma-27b-it was queried once with each of five prompt variants, indexed 0 to 4, at temperature 0.0. They are variants rather than paraphrases: each changes the instruction, not only its wording (texts in the Supplementary Information). Variant 0 is the unmodified V3 prompt. Variant 1 prepends an instruction to attend to architecture. Variant 2 is built from V4 rather than V3, imposes step-by-step reasoning, lists the nine classes alphabetically and states disambiguation rules. Variant 3 adds the clinical context of a colorectal surgical specimen, and variant 4 instructs the model to lower its stated confidence under ambiguity. Variant 2 also behaves differently: its confidence-correctness AUROC on the unweighted subsets is 0.609 on validation and 0.628 externally, against 0.49 to 0.51 for the other four, and its accuracy is the lowest of the five.

\textbf{Wording-only paraphrases.} A further family, V3W, separates wording from instruction. It keeps V3's instruction content, class names and order, the content of each class cue and the output format, and rewrites only the role sentence, the list header and the wording of each cue, giving V3 itself and four paraphrases (texts in the Supplementary Information). It ran on the same 1,800 validation patches on 6 October 2026, between 03:30 and 08:03 UTC, and on the same 1,800 external patches on 17 October 2026, between 06:39 and 08:51 UTC. Its variant 0 is byte-identical to V3, so it also measures label drift between that session and the stored run.

\textbf{What agreement depends on.} Every patch receives the same five prompts, so the image is the only input that varies between patches: with the image removed, agreement could not vary from patch to patch at all. Three controls test this and the narrower question of which property of the image agreement follows. Each variant was sent ten times with no image and ten times with a uniform grey image, recording the answers the prompts alone produce. All 1,800 validation patches were sent under the five variants with their pixels randomly permuted, which keeps each patch's colour histogram and removes all tissue morphology. And 360 patches, 40 per class, were each sent with the image of a patch from a different class, to check that the answers follow the image supplied. The controls ran on 6 October 2026 alongside the V3W run: the no-image and grey calls between 03:28 and 03:30 UTC, the swapped images between 03:29 and 04:20, and the permuted images between 04:20 and 08:30.

\textbf{The single-query baseline is variant 0.} Variant 0 is the V3 prompt issued inside the same session as the other four, so every comparison in Section~\ref{subsec:consistency-results} is in-session by construction. Session identity matters more here than it would for locally hosted weights. Variant 0 and the stored full-scale V3 prediction disagree on 54 of 1,800 validation patches (3.0\%) and on 0 of 1,800 external-test patches, and no single per-call disagreement rate produces both, so those two runs did not meet the same endpoint conditions. Fifty-four drifted labels is the same order as the discordant counts that drive the voting test, so a cross-session baseline would measure part of that drift rather than the effect of voting.

Every call in this paper goes to one OpenAI-compatible endpoint, \texttt{api.ai.it.ufl.edu}, a University of Florida gateway. Request timestamps are logged for the log-probability runs of Section~\ref{subsec:logprob-methods}, which ran on 1 September 2026 between 22:07 and 23:53 UTC, and for the V3W and control runs. For the zero-shot and K=5 runs the closest available inference date is when their output files were written, 25 June to 3 July 2026. The endpoint exposes no model snapshot or version string.

For each patch we compute: (1) the modal label across K queries; (2) consistency score $c = (\text{queries agreeing with modal}) / K$; and (3) uncertainty score $1 - c$. An unknown answer (Section~\ref{subsec:models}) takes no part in the vote but stays in the denominator, so it counts as dissent; a patch with no answer naming a listed class has no modal label, scores $c = 0$ and counts as an error, and averaged confidence includes the stated confidence of every answer, unknown ones included. We also compute the predictive entropy $H = -\sum_\ell p_\ell \log_2 p_\ell$ over the empirical label distribution $\{p_\ell\}$ across the K queries, routing by $-H$, and report it as a reparameterization of the consistency score rather than as an independent signal. At K=5 the attainable entropies are 0.000, 0.722, $\{0.971, 1.371\}$, $\{1.522, 1.922\}$ and 2.322 bits, and those ranges do not overlap across consecutive modal counts, so $-H$ orders patches by modal fraction and cannot reorder two that differ in it. What it adds is a split inside the 3/5 and 2/5 levels, where dissent concentrated on one alternative label scores differently from dissent spread across two. That makes it an order-preserving refinement, visible in Table~\ref{tab:t6b} as seven distinct values against five and a lower tie fraction, and it still uses only how the K labels partition rather than which labels they are.

A signal that is genuinely not a function of the modal count has to use which labels the dissenting responses chose. \textbf{Confusability-weighted agreement} credits each response by how confusable its label is with the modal label, so a patch splitting 3/2 toward a neighbouring tissue class scores higher than one splitting 3/2 toward an unrelated class. Confusability comes from MedGemma's own full-scale validation confusion matrix, symmetrized, row-normalized and scaled so the most confusable pair carries weight 1, so no test label enters the score. For the validation evaluation the matrix is estimated from the 8,204 validation patches outside the 1,800, so the evaluated patches do not inform their own weights. A uniform weight matrix reduces the score to plain consistency and serves as a control.

Three kinds of comparison are made. Voting is assessed as the cohort-level accuracy difference between the modal vote and the in-session single query. Routing signals are compared by paired differences in selective-accuracy AUC; each signal routes its own outcome, the modal vote for agreement-based signals and the single query for single-query confidence, and because a contrast between the two then changes signal and outcome together, it is repeated with both signals routing one fixed outcome, first the modal-vote correctness and then the single-query correctness. Finally, agreement is recomputed with each variant left out in turn and with the three V3 edits alone (variants 0, 1 and 3), to see which variants carry the signal, and over every subset of $K = 2$ to 5 prompts within each family and of up to nine prompts pooling the V3 variants with the four V3W paraphrases, which ran on the same validation patches. Each variant's own stated confidence is also evaluated as a single-query signal routing that variant's answers.

\subsection{Label-Token Confidence}\label{subsec:logprob-methods}

Everything above measures the integer the model writes in its response. A second quantity is available from the same call: the probability the model assigns to the label token itself. Comparing the two is meant to separate a failure of verbalization from a failure of the underlying representation, a separation the renormalization described below limits.

We keep the V3 prompt byte-identical and request token log-probabilities with the top 20 alternatives at each position. At the first token of the label span, the token immediately after \texttt{LABEL: }, we map each returned alternative to a class by prefix and renormalize over the nine classes. The probability of the predicted class under that renormalization is the label-token confidence. Because it comes from the same response as the verbalized number, the two are paired on identical patches by construction, so any difference between them isolates the confidence source rather than the sample.

Two alternative designs were rejected. Constrained single-token decoding would change the elicitation and break comparability with the verbalized number, and scoring nine separate completions per patch would cost nine times as much while measuring the model's preference among forced choices rather than the confidence attached to the answer it gave.

The design rests on the first label token identifying the class. A 50-patch pilot per model confirms it: the first tokens observed are \texttt{adipose}, \texttt{background}, \texttt{cancer}, \texttt{colorectal}, \texttt{debris}, \texttt{lymphocytes}, \texttt{normal} and \texttt{smooth}, with no collisions, and the label span was located in all 100 pilot responses. The renormalization covers only the classes that appear among the 20 alternatives: on average 8.22 of nine for MedGemma on validation and 8.25 externally, and 6.71 and 6.75 for Gemma-3. All nine appear for 937 of MedGemma's 2,001 validation patches and 1,030 of its 2,000 external patches, against 370 and 389 for Gemma-3. A class outside the top 20 adds nothing to the denominator, and the stored outputs keep only the renormalized value and the class count, so the missing probability mass cannot be bounded. Every label-token statistic is therefore also computed on the patches where all nine classes appeared, and within strata of the class count.

Sampling here is fixed and differs from the tables above, so these rows carry their own $n$. Validation reuses the 2,001-patch held-out evaluation partition, which keeps the calibration figures comparable to Table~\ref{tab:t2}. The external split is a fresh unstratified random sample of 2,000 patches at seed 42; reusing the 1,800-patch consistency subset would bias the accuracy base, since it is stratified on correctness. Both models ran on both samples, 8,002 calls in all, with no call failures and the label span located in every response.

\section{Results}\label{sec:results}

\subsection{In-Distribution Classification}\label{subsec:indist}

Table~\ref{tab:t1} shows patch-level results on all 10,004 validation patches.

\begin{table}[h]
\caption{Validation set classification (NCT-CRC-HE-100K, $n = 10{,}004$). Confidence intervals are Wilson score intervals on accuracy. Macro F1 weights all nine classes equally; weighted F1 weights by class support and is therefore less dominated by the near-zero classes (Table~\ref{tab:perclass}). The ResNet-18 row is leakage-inflated (Section~\ref{subsec:models}).}\label{tab:t1}
\begin{tabular}{@{}lcccc@{}}
\toprule
Model & Accuracy & 95\% Wilson CI & Macro F1 & Weighted F1 \\
\midrule
MedGemma-27b-it V3 (zero-shot) & 0.4063 & 0.397--0.416 & 0.3272 & 0.3455 \\
Gemma-3-27b-it V3 (zero-shot) & 0.3548 & 0.345--0.364 & 0.2702 & 0.2972 \\
ResNet-18 64~px (supervised) & 0.9971 & 0.996--0.998 & 0.9971 & 0.9971 \\
\botrule
\end{tabular}
\end{table}

MedGemma outperforms Gemma-3 by 5.2~pp in accuracy and 5.7~pp in macro F1. The 95\% CI half-widths at $n = 10{,}004$ are approximately $\pm 0.010$, making the 5.2~pp gap statistically clear. Both VLMs clear the 11.1\% uniform rate comfortably, but the uniform rate is not the right reference for an unbalanced split: against the majority class (TUM, 14.3\% on validation) MedGemma is 26.3~pp above and Gemma-3 21.2~pp above. On the external test split the majority class is ADI at 18.6\%, so the margins there are 16.6~pp and 12.3~pp.

\begin{figure}[H]
\centering
\includegraphics[width=0.85\textwidth]{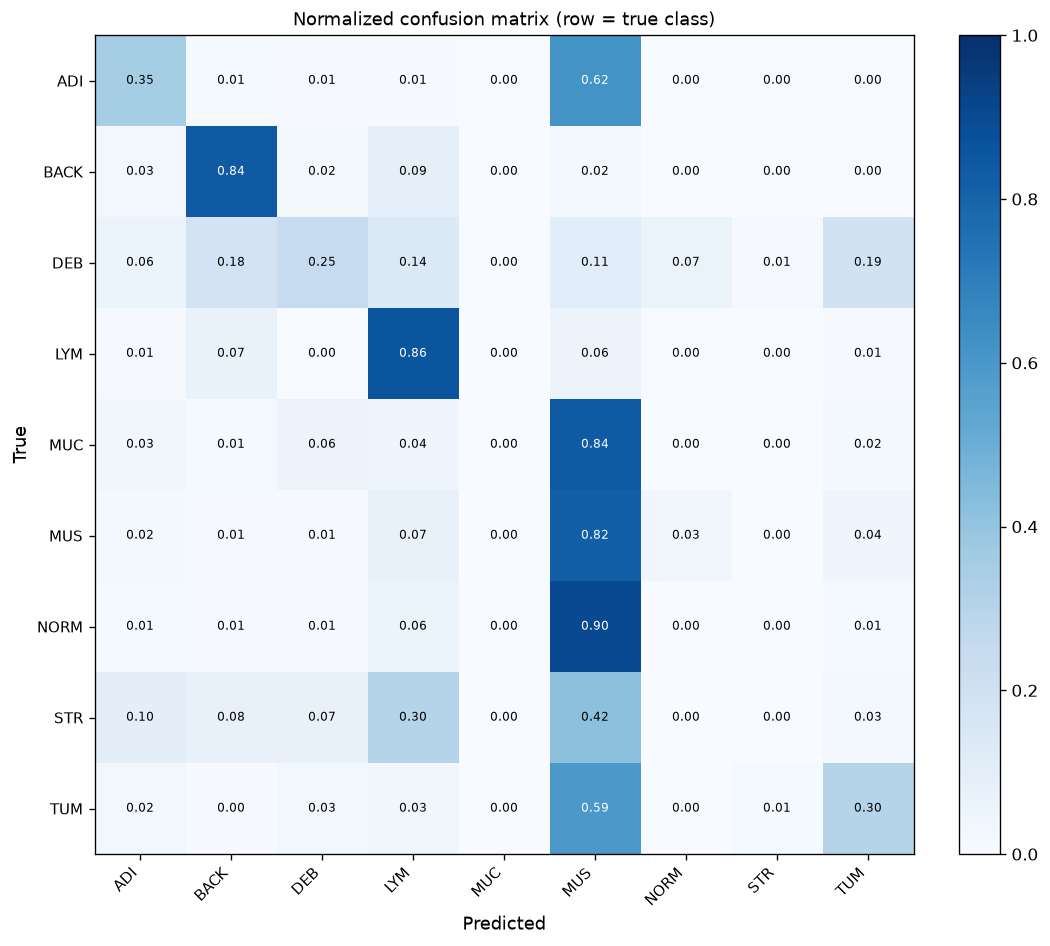}
\caption{MedGemma-27b-it normalized confusion matrix, full-scale validation set ($n=10{,}004$). Rows are true classes and columns are predicted classes; each row sums to 1. The dense MUS (smooth muscle) column, into which several true classes are misrouted, visualizes the over-prediction bias discussed in Section~\ref{subsec:disc-routing}.}\label{fig:mg-cm}
\end{figure}

\begin{figure}[H]
\centering
\includegraphics[width=0.85\textwidth]{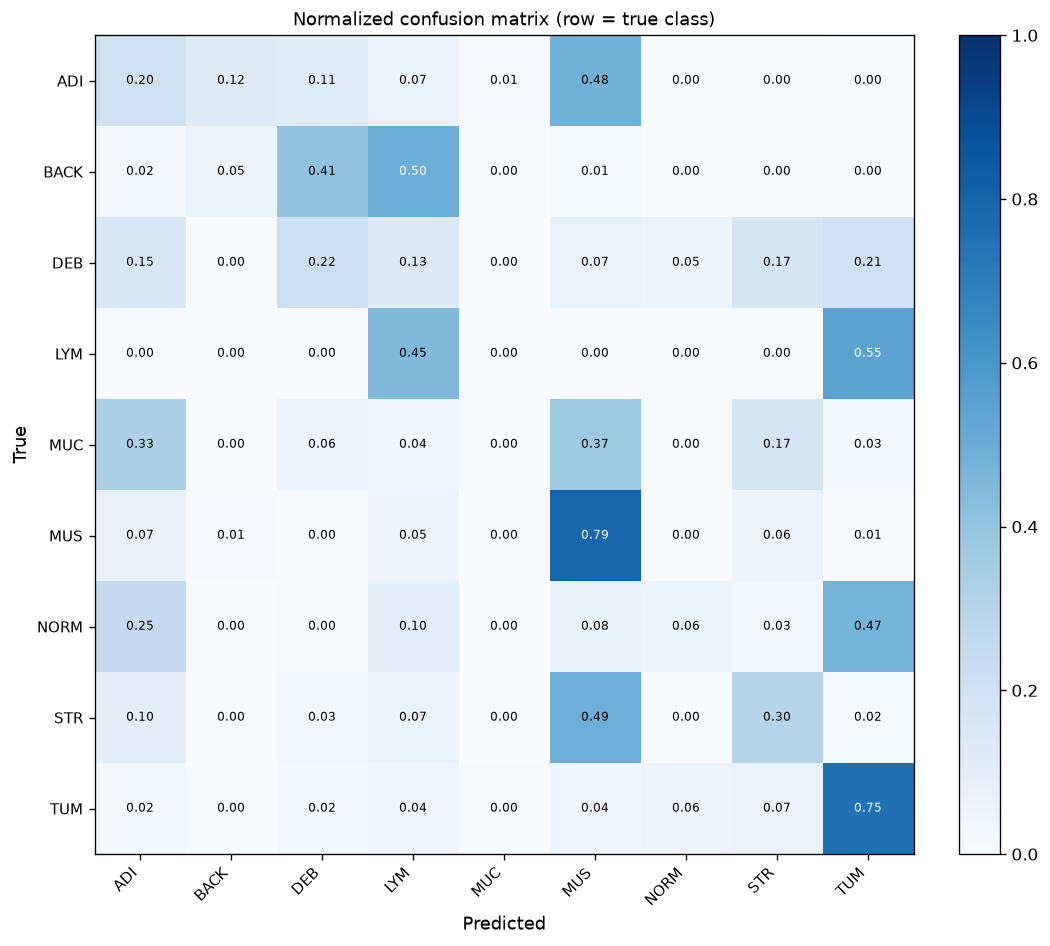}
\caption{Gemma-3-27b-it normalized confusion matrix, full-scale validation set ($n=10{,}004$). Rows are true classes and columns are predicted classes; each row sums to 1. The error pattern is complementary to MedGemma's (Figure~\ref{fig:mg-cm}): TUM and MUS are the dominant predicted classes, while BACK and ADI are frequently misclassified.}\label{fig:g3-cm}
\end{figure}

The per-class breakdown exposes the failure modes (Figures~\ref{fig:mg-cm} and \ref{fig:g3-cm}). MedGemma achieves solid F1 on BACK (0.75), LYM (0.67), ADI (0.42), and TUM (0.39), but near-zero on STR (0.00), NORM (0.004), and MUC (0.002). Smooth muscle remains over-predicted: MUS recall is high (0.82) with precision of only 0.23, meaning most MUS predictions are wrong. Gemma-3 distributes its predictions differently. TUM (0.56) and MUS (0.54) are its strongest classes, and STR (0.33) and NORM (0.10) are non-zero, but BACK (0.09) and ADI (0.19) collapse. Section~\ref{subsec:disc-tuning} takes up whether these complementary patterns explain the routing results; they do not.

Table~\ref{tab:perclass} gives the full per-class breakdown. The macro F1 is held down by three classes on which MedGemma is at or near zero (MUC, NORM, STR), so it understates how the model performs on the classes that dominate the split; the weighted F1 (Table~\ref{tab:t1}) is correspondingly higher.

\begin{table}[h]
\caption{Per-class F1 for both VLMs on validation (val) and external test (test). Classes are the nine NCT-CRC tissue types; the two right-most rows repeat the macro and support-weighted averages. Near-zero cells (MUC, NORM, STR for MedGemma) are the ones that pull down macro F1.}\label{tab:perclass}
\begin{tabular}{@{}lcccc@{}}
\toprule
 & \multicolumn{2}{c}{MedGemma-27b-it} & \multicolumn{2}{c}{Gemma-3-27b-it} \\
\cmidrule(lr){2-3}\cmidrule(lr){4-5}
Class & val & test & val & test \\
\midrule
Adipose (ADI) & 0.425 & 0.472 & 0.188 & 0.200 \\
Background (BACK) & 0.750 & 0.833 & 0.091 & 0.108 \\
Debris (DEB) & 0.344 & 0.077 & 0.242 & 0.060 \\
Lymphocytes (LYM) & 0.669 & 0.719 & 0.379 & 0.348 \\
Mucus (MUC) & 0.002 & 0.000 & 0.009 & 0.002 \\
Smooth muscle (MUS) & 0.360 & 0.216 & 0.538 & 0.371 \\
Normal mucosa (NORM) & 0.004 & 0.008 & 0.101 & 0.253 \\
Stroma (STR) & 0.000 & 0.009 & 0.326 & 0.285 \\
Tumor epithelium (TUM) & 0.390 & 0.322 & 0.559 & 0.589 \\
\midrule
Macro F1 & 0.327 & 0.295 & 0.270 & 0.246 \\
Weighted F1 & 0.346 & 0.328 & 0.297 & 0.258 \\
\botrule
\end{tabular}
\end{table}

\subsection{Calibration}\label{subsec:calibration-results}

Table~\ref{tab:t2} shows calibration metrics on the held-out evaluation partition.

\begin{table}[h]
\caption{Calibration on held-out evaluation partition ($n_\text{cal} \approx 8{,}000$, $n_\text{eval} = 2{,}001$). ECE (raw) is equal-width 15-bin; ECE (adapt.) is 15 equal-mass bins; Brier is the binning-free raw Brier score.}\label{tab:t2}
\begin{tabular}{@{}lccccc@{}}
\toprule
Model & ECE (raw) & ECE (adapt.) & Brier & T & ECE (calibrated) \\
\midrule
MedGemma-27b-it & 0.553 & 0.553 & 0.546 & 200 & 0.106 \\
Gemma-3-27b-it & 0.535 & 0.535 & 0.497 & 200 & 0.146 \\
ResNet-18 64~px & 0.003 & 0.001 & 0.003 & 1.12 & 0.003 \\
\botrule
\end{tabular}
\end{table}

\begin{figure}[H]
\centering
\includegraphics[width=0.95\textwidth]{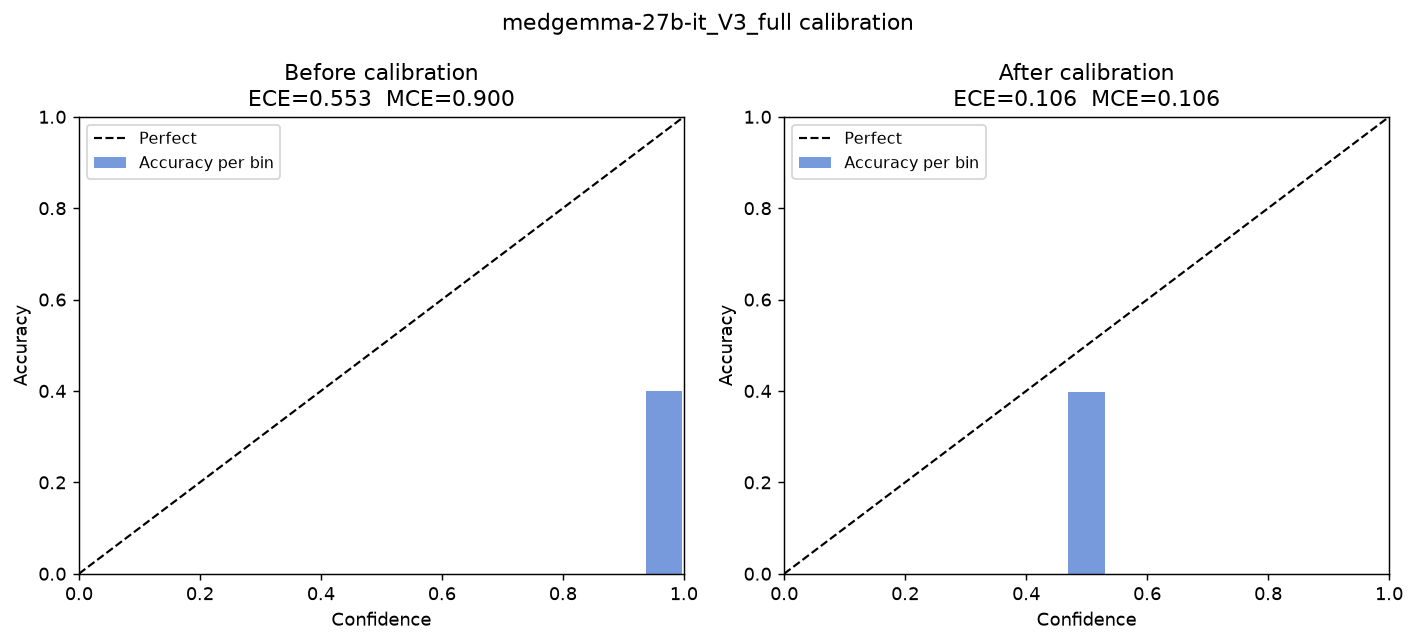}
\caption{MedGemma-27b-it reliability diagrams before (left, ECE = 0.553) and after (right, ECE = 0.106) temperature scaling, held-out evaluation partition. Responses cluster near confidence 0.9--1.0 regardless of actual accuracy ($\approx$0.40); the fitted temperature $T=200$ collapses calibrated confidence toward 0.5 for nearly all inputs.}\label{fig:mg-rel}
\end{figure}

\begin{figure}[H]
\centering
\includegraphics[width=0.95\textwidth]{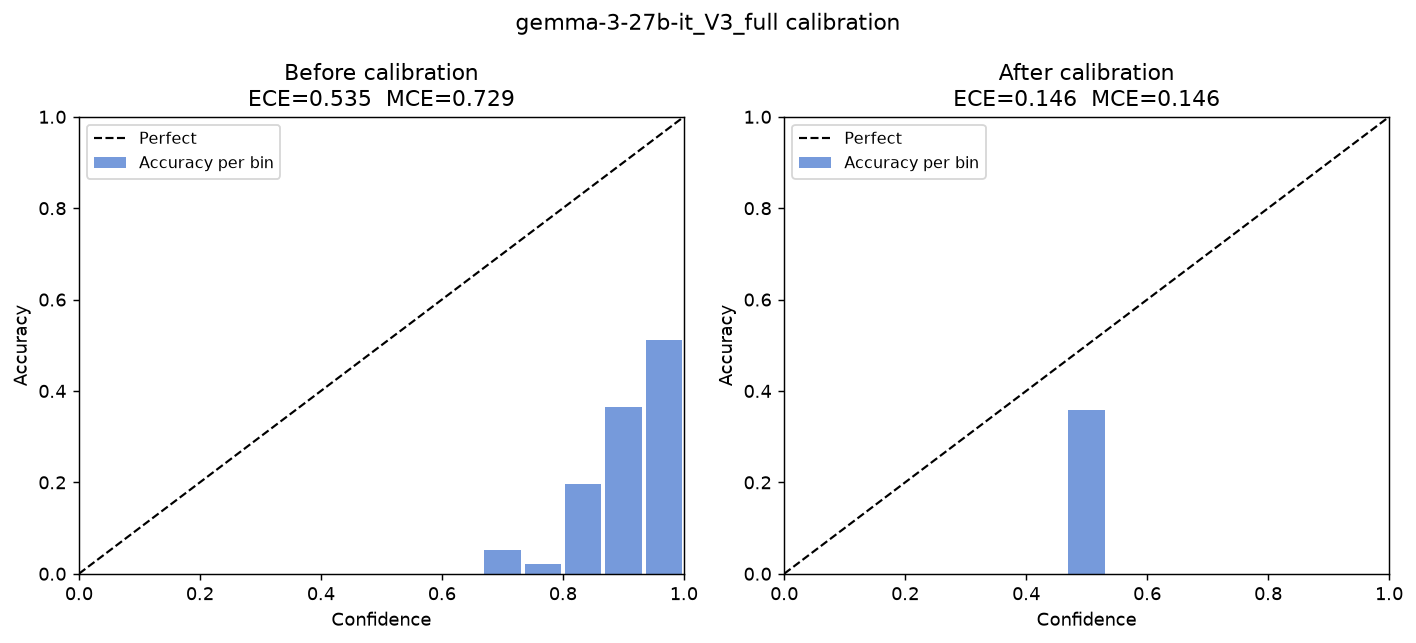}
\caption{Gemma-3-27b-it reliability diagrams before (left, ECE = 0.535) and after (right, ECE = 0.146) temperature scaling, held-out evaluation partition. The same T=200 saturation pattern occurs as for MedGemma (Figure~\ref{fig:mg-rel}), though the raw confidence distribution before calibration spans a wider bin range.}\label{fig:g3-rel}
\end{figure}

Both VLMs produce ECE above 0.53, with responses clustered in the 70--95\% confidence range while actual accuracy is 40.6\% for MedGemma and 35.5\% for Gemma-3 (Figures~\ref{fig:mg-rel} and \ref{fig:g3-rel}). Adaptive (equal-mass) ECE and the Brier score tell the same story (Table~\ref{tab:t2}), so the miscalibration is not a binning artifact. Temperature scaling reduces ECE to 0.106 and 0.146 respectively, but the fitted T of 200 is exactly the upper bound of the search interval (Section~\ref{subsec:calibration-methods}) for both models, so it is where the optimizer stopped rather than an estimated quantity. At that temperature the calibrated confidences all sit within 0.02 of 0.5, and the calibrated ECE therefore reduces to $|0.5 - \mathrm{accuracy}|$: 0.102 against the reported 0.106 for MedGemma and 0.143 against 0.146 for Gemma-3, with the residual explained by that remaining deviation from 0.5. The apparent ECE improvement says only that predicting 0.5 uniformly is less wrong than predicting 70--95\% uniformly. We draw no calibration conclusion from it.

An intercept removes the saturation without supplying information. Platt scaling with a slope and an intercept, and isotonic regression, bring MedGemma's held-out ECE to 0.010 and 0.011, but their Brier scores of 0.240 and 0.239 are those of assigning the calibration split's accuracy to every patch (0.240): the fitted Platt slope is $-0.017$, isotonic regression maps every stated confidence from 95 to 100 to 0.41, and the resolution term of the Brier score is 0.0006. Externally the two maps reach Brier 0.233 and 0.230 against 0.231 for the constant. Gemma-3 has something to calibrate. Isotonic regression lowers its held-out Brier score to 0.200 against 0.230 for the constant, with resolution 0.030, and to 0.186 against 0.216 externally (Supplementary Information). The temperature saturates because of the confidences, and a less restricted map lowers MedGemma's calibration error without changing what its confidence says about correctness.

A saturating temperature says only that monotone scaling cannot help. How much information the confidence carries is a separate question, and the first thing to report is how little there is to rank. MedGemma emits four distinct confidence values on validation and five on the external test split, with 96\% of patches at a single value, so 92\% of patch pairs are tied. Gemma-3 uses nine values with a tie fraction of 0.33 (0.30 on test). With ties handled by mid-ranks, MedGemma's confidence-correctness AUROC is 0.499 (95\% bootstrap CI [0.495, 0.503]) on validation and 0.528 ([0.523, 0.533]) on the external test split; Gemma-3 reaches 0.698 ([0.689, 0.706]) and 0.709 ([0.699, 0.720]).

Mutual information tells the same story once the estimator is pinned down. We use the plug-in estimator on 15 equal-width bins over $[0, 1]$, the same binning as the ECE, and compare each value against a permutation null obtained by shuffling correctness 1,000 times with the confidence distribution held fixed. This matters because the plug-in estimator is positively biased, so a small positive value is not by itself evidence of dependence. MedGemma's confidence occupies only 2 of the 15 bins on validation and yields 0.0017 bits against a null mean of 0.0001 bits; on the 2,001-patch evaluation partition it yields 0.0015 bits against a null mean of 0.0005 bits, which the permutation test does not distinguish from chance ($p = 0.17$). Gemma-3 occupies 8 bins and yields 0.111 bits on validation and 0.111 on test, both far above their nulls ($p < 0.002$). So MedGemma's confidence is at or near the estimator's floor at the sample sizes we have, and Gemma-3's is not.

MedGemma's confidence is not anti-informative; it is close to constant, which leaves nothing for a threshold to separate and nothing for temperature scaling to reorder. Gemma-3's confidence spans a usable range and carries a modest signal that survives temperature scaling, since scaling is monotone and preserves ranking. Verbalized VLM confidence is therefore uninformative for MedGemma under this prompt and not for Gemma-3.

The ResNet-18 row of Table~\ref{tab:t2} comes from leakage-inflated validation patches and says nothing about in-distribution calibration; the CNN's calibration on the external cohort is in Section~\ref{subsec:crosscenter}.

\subsection{Routing}\label{subsec:routing-results}

Table~\ref{tab:t3} shows routing behavior across both models and both evaluation sets: the selective-accuracy AUC $A_{\text{AUC}}$ (Equation~\ref{eq:aauc}, higher is better), its paired-bootstrap gap over random routing with a 95\% CI, and the auto-confirm accuracy at the $b = 0.15$ operating point.

\begin{table}[h]
\caption{Routing by confidence. $A_{\text{AUC}}$ (cal) is the selective-accuracy AUC of calibrated confidence routing with ties resolved by their expectation over random tie orders; $A_{\text{AUC}}$ (rnd) is random routing, whose expected auto-confirm accuracy equals the base accuracy at every budget, so it is exactly the accuracy in Table~\ref{tab:t4} and is the reference for the gap column. The gap column is the point estimate with its 95\% paired-bootstrap CI; a CI excluding zero (marked $^\ast$) means the signal is distinguishable from random under the independence assumption, which Table~S1 then relaxes. ``Prior'' is a baseline that ranks patches only by the validation accuracy of the class the model predicted, making no use of the confidence value; on the validation rows it is estimated in-sample and therefore optimistic. ``sd'' is the standard deviation of $A_{\text{AUC}}$ (cal) that the tie ordering alone induces over 200 random tie orders, and ``Ties'' the fraction of patch pairs sharing a confidence value. ``Cal @15\%'' is auto-confirm accuracy at a 15\% specialist budget.}\label{tab:t3}
\footnotesize
\setlength{\tabcolsep}{2pt}
\begin{tabular}{@{}llccccccc@{}}
\toprule
Model & Set & $A_{\text{AUC}}$ (cal) & sd & $A_{\text{AUC}}$ (rnd) & Gap [95\% CI] & Prior & Ties & Cal @15\% \\
\midrule
MedGemma-27b-it & val & 0.402 & 0.004 & 0.406 & $-0.004$ [$-0.010$, $0.002$] & 0.554 & 0.92 & 0.407 \\
MedGemma-27b-it & test & 0.389 & 0.004 & 0.352 & $+0.037$ [$0.029$, $0.044$]$^\ast$ & 0.577 & 0.92 & 0.357 \\
Gemma-3-27b-it & val & 0.469 & 0.003 & 0.355 & $+0.114$ [$0.108$, $0.120$]$^\ast$ & 0.416 & 0.33 & 0.404 \\
Gemma-3-27b-it & test & 0.425 & 0.004 & 0.309 & $+0.115$ [$0.108$, $0.123$]$^\ast$ & 0.392 & 0.30 & 0.356 \\
ResNet-18 64~px & val & 0.9998 & 0.000 & 0.9971 & $+0.003$ [$0.002$, $0.004$]$^\ast$ & 0.999 & 0.00 & 0.9996 \\
ResNet-18 64~px & test & 0.978 & 0.000 & 0.911 & $+0.067$ [$0.061$, $0.073$]$^\ast$ & 0.895 & 0.00 & 0.963 \\
\botrule
\end{tabular}
\end{table}

\begin{figure}[H]
\centering
\includegraphics[width=0.95\textwidth]{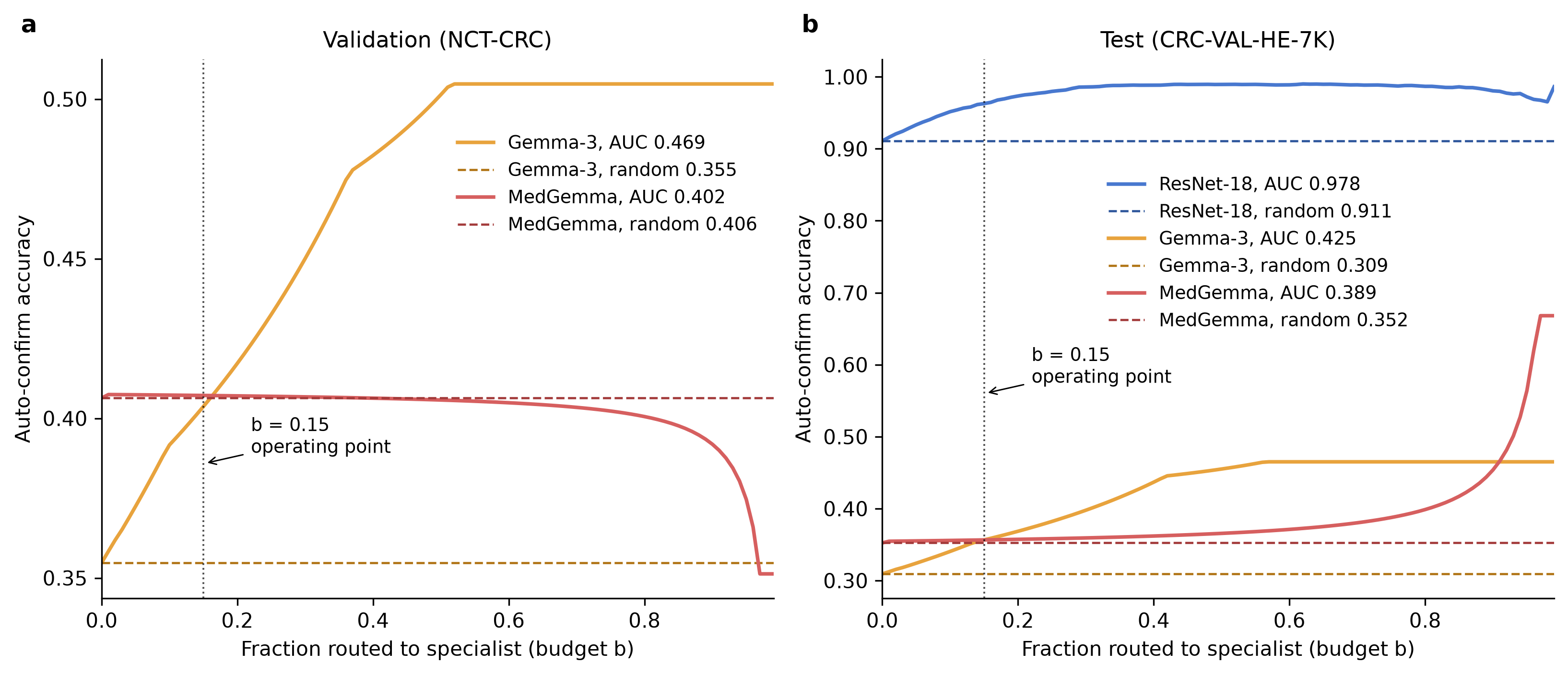}
\caption{Selective-accuracy routing curves (auto-confirm accuracy versus fraction routed) for calibrated confidence, full-scale validation (left) and external test (right); the dotted vertical line marks the $b=0.15$ operating point used in Table~\ref{tab:t3}. Curve endpoints match Table~\ref{tab:t3} to three decimals. Each dashed line is the random-routing reference for the model of the same colour; random routing has the model's base accuracy at every budget, so the line is flat at that accuracy. ResNet-18 64~px (blue) is drawn on the external panel only, because its validation patches share slides with its training patches. It routes far above its reference, Gemma-3 (orange) routes above its own on both splits, and MedGemma (red) tracks its own reference closely, falling marginally below on validation and sitting marginally above on external test.}\label{fig:routing}
\end{figure}

The supervised reference routes well on the external cohort (Figure~\ref{fig:routing}): at a 15\% budget its auto-confirm accuracy is 96.3\% against 91.1\% for random, and its selective-accuracy AUC gap over random is $+0.067$ (95\% CI [0.061, 0.073]). Its validation rows are leakage-inflated and are not interpreted.

Zero-shot PLIP, a pathology image-text encoder used without any task training, is more accurate than both VLMs, and its confidence ranks better than MedGemma's but not as well as Gemma-3's. It reaches 49.7\% accuracy on validation and 53.8\% on the external cohort, against 40.6\% and 35.2\% for MedGemma and 35.5\% and 30.9\% for Gemma-3. Its confidence-correctness AUROC is 0.586 [0.576, 0.597] and 0.593 [0.580, 0.606], with routing gaps over random of $+0.090$ [0.082, 0.098] and $+0.098$ [0.088, 0.107], against 0.698 and 0.709 and $+0.114$ and $+0.115$ for Gemma-3. A linear probe on its frozen image features reaches 94.7\% external accuracy, with a routing gap of $+0.044$ [0.039, 0.048]. The patches therefore carry enough information both to classify them and to rank the classifications. The interface does not decide whether a confidence ranks: Gemma-3's generated number ranks better than PLIP's four-template score, while PLIP with a single template ranks better than Gemma-3 on validation (AUROC 0.708, gap $+0.171$; Supplementary Information). QuiltNet, a second pathology image-text model run under the same templates, gives the same picture: 52.7\% and 50.3\% accurate, AUROC 0.672 [0.663, 0.683] and 0.684 [0.671, 0.696], and routing gaps of $+0.148$ [0.141, 0.156] and $+0.165$ [0.156, 0.173], larger than Gemma-3's although its AUROC is lower, with a linear probe at 95.4\% external accuracy. This does not show that MedGemma's own image encoder carries the same information, and PLIP's confidence is not calibrated either (ECE 0.137 and 0.121).

MedGemma's verbalized confidence does not provide a usable routing signal. On validation the gap over random is $-0.004$ (95\% CI [$-0.010$, $0.002$]), indistinguishable from zero. On the external test split it is $+0.037$ ([0.029, 0.044]), a third of Gemma-3's. A signal worth nothing on one cohort and 0.04 of selective-accuracy AUC on the other is not one a routing system can use. MedGemma's confidence is also beaten decisively on both splits by a baseline that ignores the confidence value entirely and ranks patches by the validation accuracy of the class the model predicted (0.554 against 0.402 on validation, 0.577 against 0.389 on test). Whatever ordering information exists in a MedGemma prediction lives in the label, not in the number attached to it.

Gemma-3's confidence does route. Its gap over random is $+0.114$ on validation and $+0.115$ externally (95\% CI [0.108, 0.120] and [0.108, 0.123]), and at 15\% budget calibrated routing lifts auto-confirm accuracy from 35.5\% to 40.4\% on validation. It also beats the predicted-class-prior baseline on both splits (0.469 against 0.416, 0.425 against 0.392), so unlike MedGemma it is contributing something beyond class identity. A gap of 0.11 is modest: useful, stable across cohorts, and far below the CNN.

\textbf{Confidence by predicted class.} MedGemma's mean confidence barely varies with the class it predicts: 0.9499 on MUS, 0.9499 on LYM, 0.9500 on TUM and ADI, and 0.9637 on BACK, the five classes it predicts more than 600 times, whose accuracies range from 0.23 to 0.68. Across all nine predicted classes the spread is 0.016, the lowest mean being 0.9482 on STR, which it predicts 28 times. Within a predicted class the confidence is usually a single value, so the within-class AUROC is 0.50 to two decimals on MUS, LYM, TUM and ADI, which together take 80\% of all predictions. The two classes where it does vary point in opposite directions under a 2,000-resample bootstrap: background (three values, sd 0.022, $n = 1{,}307$) gives 0.299 [0.271, 0.325], below chance, and debris (two values, sd 0.009, $n = 498$) gives 0.533 [0.516, 0.551], slightly above it. Where the signal moves at all, it has no consistent direction. Gemma-3 differs on exactly this axis. Its per-class mean confidence spans 0.199 (0.751 on NORM to 0.950 on BACK) and its within-class AUROC stays high (0.75 on MUS, 0.96 on LYM, 0.85 on DEB), so its signal is not merely a restatement of which class it picked. Section~\ref{subsec:disc-routing} takes up what this implies about the mechanism.

\textbf{Sensitivity to clustering.} Every interval above treats patches as independent, and they are not. Table~S1 recomputes each headline interval under a design effect and reports the break-even intra-cluster correlation, the $\rho$ at which an interval stops excluding zero. MedGemma's validation gap includes zero before any correction, and its external-test gap goes at $\rho = 0.09$, below what patches sharing a slide plausibly reach. Gemma-3's routing gain survives throughout, needing $\rho$ above 0.85 on the external split and beyond 1 on validation. The consistency and label-token contrasts of Sections~\ref{subsec:consistency-results} and~\ref{subsec:logprob-results} are where the cluster unit bites hardest, and those sections give their break-even values.

\subsection{Cross-Cohort Generalization}\label{subsec:crosscenter}

Neither VLM was trained on either cohort, so the change in its performance from the validation split to the external split measures the cohort-and-batch shift as that model sees it (Table~\ref{tab:t4}).

\begin{table}[h]
\caption{VLM performance on validation (NCT-CRC-HE-100K) and the external cohort (CRC-VAL-HE-7K). ``Rel.'' is the drop divided by validation accuracy. ``Err.\ ratio'' is the external error rate divided by the validation error rate. ``Retained'' is the fraction of above-majority-class accuracy retained, using each split's own majority-class rate (Table~\ref{tab:classdist}).}\label{tab:t4}
\footnotesize
\setlength{\tabcolsep}{3.5pt}
\begin{tabular}{@{}lccccccccc@{}}
\toprule
Model & Val Acc & Test Acc & $\Delta$ Acc & Rel. & Err.\ ratio & Retained & Val F1 & Test F1 & $\Delta$ F1 \\
\midrule
MedGemma-27b-it & 0.4063 & 0.3524 & $-0.054$ & 0.133 & 1.09 & 0.63 & 0.3272 & 0.2950 & $-0.032$ \\
Gemma-3-27b-it & 0.3548 & 0.3093 & $-0.045$ & 0.128 & 1.07 & 0.58 & 0.2702 & 0.2461 & $-0.024$ \\
\botrule
\end{tabular}
\end{table}

MedGemma loses 5.4~pp of accuracy and Gemma-3 4.5~pp, with macro F1 falling by 0.032 and 0.024, so MedGemma's lead narrows from 5.2~pp to 4.3~pp. How large those drops look depends on the scale. The error rates barely move, rising by factors of 1.09 and 1.07, because at 35 to 41\% accuracy most patches are already misclassified. Measured against the headroom above each split's majority class, the same drops are large: the two models keep 0.63 and 0.58 of their above-majority accuracy.

The supervised reference reaches 0.911 accuracy and 0.884 macro F1 on the external cohort. Its drop from validation cannot be estimated, because its validation accuracy is leakage-inflated (Section~\ref{subsec:models}), so we make no comparison of robustness between the VLMs and the CNN.

Calibration on the external cohort, with the validation-fitted temperature applied unchanged, separates the two kinds of model. For the VLMs it carries no calibration information. At T = 200 the calibrated confidence is 0.5 for every patch to within 0.02, so calibrated ECE equals $|0.5 - \mathrm{accuracy}|$ and moves only with accuracy: 0.152 for MedGemma and 0.193 for Gemma-3 externally, changes of $+0.0459$ and $+0.0476$ from the 2,001-patch evaluation partition against accuracy drops of $0.0459$ and $0.0477$ over the same patches. The CNN's external ECE is a real measurement on patients it never saw: 0.058 raw and 0.054 after its validation-fitted temperature, so a single monotone rescaling fitted on the source collection corrects almost none of its external miscalibration.

\subsection{Consistency-Based Routing}\label{subsec:consistency-results}

The K=5 subsets hold up to 100 correct and 100 incorrect full-scale predictions per class, so an average over them describes the subset. Tables~\ref{tab:t6b} to~\ref{tab:t6d} lead with cohort estimates, weighted to the whole split as described in Section~\ref{subsec:consistency-methods}, and give the unweighted subset value beside each. Every comparison is paired on the same patches.

\begin{table}[h]
\caption{Selective-accuracy AUC by routing signal (higher is better), weighted to the whole split (cohort) and unweighted on the 1,800-patch subset, with ties resolved by their expectation over random tie orders. Each signal is compared with the random row on the outcome it routes, and random routing scores the accuracy of that outcome. Ties (the fraction of tied patch pairs) and the number of distinct values are counted on the subsets, validation before external. Confusability-weighted agreement uses weights from MedGemma's own validation confusions (Section~\ref{subsec:consistency-methods}).}\label{tab:t6b}
\footnotesize
\setlength{\tabcolsep}{2pt}
\begin{tabular}{@{}llcccccc@{}}
\toprule
 & & \multicolumn{2}{c}{Validation} & \multicolumn{2}{c}{External} & & \\
\cmidrule(lr){3-4}\cmidrule(lr){5-6}
Routing signal & Outcome routed & Cohort & Subset & Cohort & Subset & Ties & Distinct \\
\midrule
Single-query confidence & Single query & 0.391 & 0.320 & 0.368 & 0.354 & 0.91 / 0.92 & 4 / 4 \\
Random & Single query & 0.400 & 0.338 & 0.342 & 0.349 & -- & -- \\
\midrule
Mean confidence, K=5 & Modal vote & 0.426 & 0.348 & 0.402 & 0.372 & 0.74 / 0.75 & 10 / 7 \\
Consistency score & Modal vote & 0.524 & 0.431 & 0.455 & 0.451 & 0.37 / 0.37 & 5 / 5 \\
Predictive entropy, K=5 & Modal vote & 0.524 & 0.431 & 0.456 & 0.451 & 0.34 / 0.35 & 7 / 7 \\
Confusability-weighted & Modal vote & 0.524 & 0.414 & 0.440 & 0.438 & 0.11 / 0.15 & 254 / 180 \\
Random & Modal vote & 0.400 & 0.341 & 0.349 & 0.357 & -- & -- \\
\botrule
\end{tabular}
\end{table}

Voting over the five variants does not raise accuracy. Weighted to the cohort, the modal vote differs from the in-session single query by $-0.000$ [$-0.006$, 0.006] on validation and $+0.007$ [$-0.002$, 0.016] externally (Table~\ref{tab:t6c}). The external subset shows a small gain, $+0.008$ [0.001, 0.016], which does not carry to the cohort.

\begin{table}[h]
\caption{Paired contrasts in selective-accuracy AUC, or in accuracy for the voting row. Cohort columns give the weighted estimate with its 95\% paired-bootstrap interval, resampled inside the selection cells; the subset column gives the unweighted point estimates, validation before external. In the two single-query rows both signals route the same outcome, the modal vote's correctness (modal) or the single query's own (own). A dagger marks the contrasts in the Holm family of Section~\ref{subsec:routing-methods}; all three hold at family-wise 0.05 on both splits.}\label{tab:t6c}
\footnotesize
\setlength{\tabcolsep}{2pt}
\begin{tabular}{@{}lccc@{}}
\toprule
Contrast & Validation cohort & External cohort & Subset \\
\midrule
Consistency $-$ random$^\dagger$ & $+0.123$ [0.105, 0.140] & $+0.106$ [0.091, 0.121] & $+0.090$ / $+0.094$ \\
Mean-of-5 conf.\ $-$ random & $+0.026$ [0.010, 0.042] & $+0.053$ [0.036, 0.069] & $+0.008$ / $+0.015$ \\
Single-query conf.\ $-$ random & $-0.009$ [$-0.022$, 0.004] & $+0.026$ [0.012, 0.041] & $-0.019$ / $+0.006$ \\
Consistency $-$ mean-of-5 conf.$^\dagger$ & $+0.098$ [0.074, 0.120] & $+0.054$ [0.031, 0.075] & $+0.083$ / $+0.079$ \\
Consistency $-$ single query (modal)$^\dagger$ & $+0.130$ [0.107, 0.153] & $+0.081$ [0.060, 0.103] & $+0.104$ / $+0.086$ \\
Consistency $-$ single query (own) & $+0.131$ [0.107, 0.154] & $+0.085$ [0.064, 0.106] & $+0.110$ / $+0.094$ \\
Confusability-weighted $-$ consistency & $+0.000$ [$-0.008$, 0.007] & $-0.015$ [$-0.022$, $-0.008$] & $-0.017$ / $-0.013$ \\
Voting gain (accuracy) & $-0.000$ [$-0.006$, 0.006] & $+0.007$ [$-0.002$, 0.016] & $+0.002$ / $+0.008$ \\
\botrule
\end{tabular}
\end{table}

Weighting changes the confidence signals from absent to weak. Single-query confidence does not separate from random on validation ($-0.009$ [$-0.022$, 0.004]) and clears it narrowly on the external cohort ($+0.026$ [0.012, 0.041]), the pattern of the full-split routing gaps in Section~\ref{subsec:routing-results}. Averaged over the five variants, confidence clears random on both splits, by $+0.026$ [0.010, 0.042] and $+0.053$ [0.036, 0.069], gains the subset did not resolve ($+0.008$ and $+0.015$, both intervals including zero) and that stop excluding zero at intra-cluster correlations of 0.08 and 0.13 (Table~S1), so we claim their direction only. On validation the averaged gain comes from variant 2, the one variant whose own confidence discriminates (Section~\ref{subsec:consistency-methods}): the mean over the other four falls below random, $-0.023$ [$-0.036$, $-0.008$]. Externally it stays above random at $+0.019$ [0.004, 0.035].

Agreement routes above every confidence signal. The consistency score clears random by $+0.123$ [0.105, 0.140] on validation and $+0.106$ [0.091, 0.121] externally. It beats averaged confidence by $+0.098$ and $+0.054$, and single-query confidence by $+0.130$ and $+0.081$ when both route the modal-vote outcome, or $+0.131$ and $+0.085$ when both route the single-query outcome, so the advantage does not come from the two signals routing different outcomes (Table~\ref{tab:t6c}). All six of these contrasts in the Holm family hold. The two external margins over confidence are the ones most exposed to clustering, breaking at intra-cluster correlations of 0.07 and 0.19 (Table~S1), so externally we claim their direction and not their size. Predictive entropy over the five labels tracks the consistency score to within 0.001 on both splits, as its construction implies (Section~\ref{subsec:consistency-methods}).

\begin{figure}[tbp]
\centering
\includegraphics[width=0.8\textwidth]{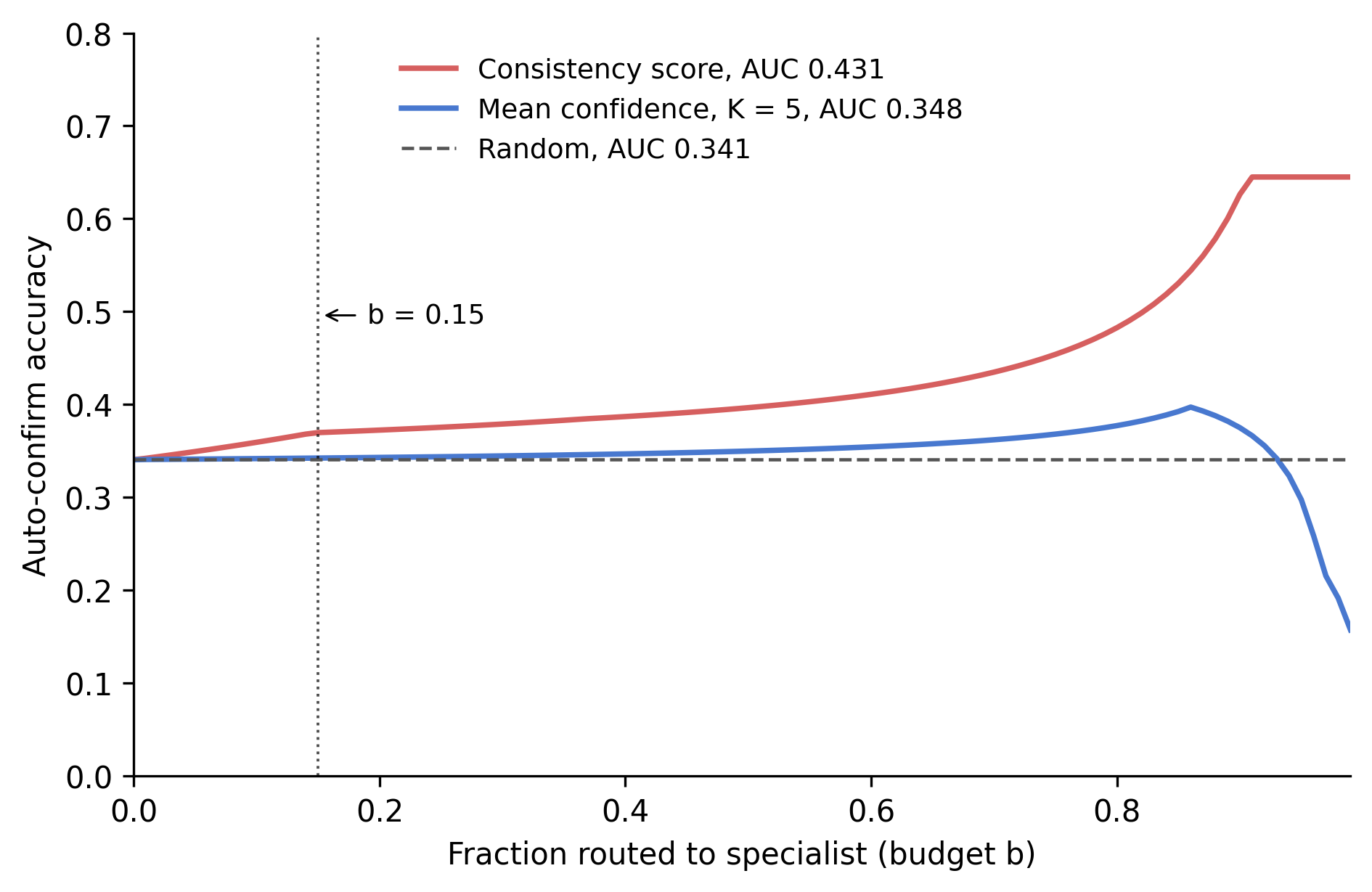}
\caption{Unweighted selective-accuracy routing curves on the 1,800-patch validation subset (V3, K=5 variants), with ties resolved by their expectation over random tie orders; the cohort-weighted values are in Table~\ref{tab:t6b}, and the legend AUCs match its subset column to three decimals. Consistency score (red) clears random (gray dashed) by a wide margin and plateaus at the 64.5\% accuracy of the 169 subset patches with 5/5 agreement; mean verbalized confidence across the five queries (blue) stays near the flat random reference (gray dashed, the subset's modal-vote accuracy) until a budget of about 0.85, then rises above it and falls below it as fewer than 270 patches remain auto-confirmed. Predictive entropy over the five labels (not shown) tracks the consistency curve to within 0.001 of AUC, being an order-preserving refinement of it that splits ties inside two of the five levels (Section~\ref{subsec:consistency-methods}).}\label{fig:consistency-routing}
\end{figure}

Agreement also sorts patches by how often the vote is right. Patches on which all five variants agree make up 10.8\% of the validation cohort and are 78.1\% correct, while those on which three or more dissent make up 13.4\% and are 17.6\% correct; on the external cohort the same strata are 11.5\% at 76.4\% and 9.5\% at 26.0\%. The validation subset holds 169 and 263 patches in these strata, at 64.5\% and 17.1\%, and Figure~\ref{fig:consistency-routing} shows its routing curves.

Most of that routing gain sits at large budgets. Routing 10, 20 or 30\% of patches by agreement raises cohort auto-confirm accuracy on validation from 40.0\% to 42.5\%, 44.2\% and 45.4\%, but externally only from 34.9\% to between 35.7\% and 35.8\%. Gemma-3's stated confidence, from one query per patch, adds 3.7 to 9.5 points at the same budgets on validation and 3.1 to 8.8 points externally (Supplementary Information). Agreement costs five queries per patch, each taking a median 0.7~s and about 510 tokens on this endpoint.

Weighting each dissent by how confusable its label is with the modal label adds nothing. With weights from MedGemma's own validation confusions, the score differs from plain consistency by $+0.000$ [$-0.008$, 0.007] on validation and falls below it externally, by $-0.015$ [$-0.022$, $-0.008$]. Weights taken from a supervised CNN's confusions on the external test split do raise the score, by $+0.024$ on validation and $+0.040$ externally, but they are built from the labels of the split being scored, so that gain is not a property of the model's answers.

\begin{table}[h]
\caption{Agreement minus random routing, cohort-weighted, with each variant left out in turn and with the three V3 edits alone, with 95\% paired-bootstrap intervals. Variant 2 is the restructured prompt built from V4 (Section~\ref{subsec:consistency-methods}); the last row is the separate wording-only family, whose answers are read from their LABEL lines while the other rows use the earlier rule (Section~\ref{subsec:models}); by the earlier rule that row is $+0.090$ [0.072, 0.107] and $+0.041$ [0.024, 0.060]. In the Holm family, the V3-edits row holds on validation and not on the external split, and the V3W row holds on both.}\label{tab:t6d}
\footnotesize
\begin{tabular}{@{}lcc@{}}
\toprule
Variants used & Validation & External \\
\midrule
All five & $+0.123$ [0.105, 0.140] & $+0.106$ [0.091, 0.121] \\
Without variant 0 & $+0.105$ [0.086, 0.122] & $+0.099$ [0.084, 0.112] \\
Without variant 1 & $+0.116$ [0.098, 0.134] & $+0.104$ [0.089, 0.118] \\
Without variant 2 & $+0.033$ [0.020, 0.045] & $-0.005$ [$-0.017$, 0.007] \\
Without variant 3 & $+0.112$ [0.094, 0.130] & $+0.116$ [0.101, 0.132] \\
Without variant 4 & $+0.145$ [0.128, 0.161] & $+0.126$ [0.112, 0.140] \\
V3 edits only (0, 1, 3) & $+0.032$ [0.021, 0.043] & $-0.010$ [$-0.021$, 0.001] \\
V3W, wording only (five) & $+0.100$ [0.084, 0.117] & $+0.054$ [0.036, 0.073] \\
\botrule
\end{tabular}
\end{table}

In the original family the agreement signal is concentrated in one variant (Table~\ref{tab:t6d}). Leaving out any of variants 0, 1, 3 or 4 keeps the gap over random between $+0.105$ and $+0.145$ on validation and between $+0.099$ and $+0.126$ externally. Leaving out variant 2 cuts it to $+0.033$ [0.020, 0.045] on validation and removes it externally, and the three V3 edits alone give $+0.032$ and $-0.010$. Within this family, agreement routes mainly because variant 2, the restructured prompt with alphabetical class order, reasoning steps and disambiguation rules, departs from the V3-style prompts on the patches the model gets wrong; edits to V3 that keep its structure carry a small signal on validation and none externally.

Adding prompts helps through variant 2 more than through their number. Averaged over every subset of the original family, agreement beats random by $+0.070$ with two prompts, $+0.090$ with three and $+0.102$ with four on validation, while subsets without variant 2 average $+0.023$ to $+0.033$ at every size. Pooling the five variants with the four V3W paraphrases, nine prompts on the same patches, gives $+0.124$, no more than the original five; the eight that exclude variant 2 give $+0.067$. Within V3W alone the gap rises from $+0.067$ for two paraphrases to $+0.100$ for five (Supplementary Information). Externally more prompts help only through variant 2: the nine pooled prompts give $+0.062$, below the original five's $+0.106$, the eight without variant 2 give $-0.017$, and the wording-only gap stays between $+0.046$ and $+0.057$ at every size.

Rewording alone also produces the signal. Under the V3W family, which keeps V3's content and changes only its wording, agreement clears random by $+0.100$ [0.084, 0.117] on the validation cohort and averaged confidence by $+0.092$ [0.064, 0.119], and leaving out any one paraphrase keeps the gap between $+0.086$ and $+0.117$, at $+0.097$ [0.081, 0.112] without paraphrase 4. The paraphrases also differ in accuracy, from 34.4\% to 45.1\% at cohort level against 40.7\% for V3 itself, the lowest being the paraphrase that on 506 of the 1,800 patches answered with a list of classes, which counts as an error. A modal vote over them gains $+0.026$ [0.013, 0.040] in accuracy where voting over the original variants gains nothing. Averaged confidence stays at random under V3W ($+0.008$ [$-0.012$, 0.029]). On the external cohort the same paraphrases give a smaller signal. Agreement clears random by $+0.054$ [0.036, 0.073], between $+0.036$ and $+0.070$ with any one paraphrase left out and $+0.065$ [0.048, 0.084] without paraphrase 4, and does not separate from averaged confidence ($+0.030$ [$-0.000$, 0.060]); the gap stops excluding zero at an intra-cluster correlation of 0.11 (Table~S1), so externally we claim its direction only. The paraphrases differ more in accuracy there, from 34.9\% to 50.2\% against 35.3\% for V3, and the modal vote gains $+0.070$ [0.050, 0.086]. Variant 0 disagrees with the stored V3 run on 30 of the 1,800 validation patches and 17 of the 1,800 external patches.

The prompts are the same for every patch, so agreement can vary between patches only through the image. With the image withheld or replaced by uniform grey, each variant gave the same label and confidence on all ten repeats. Under grey every variant answered background at confidence 1.0; with no image four variants answered normal colon mucosa and variant 1 colorectal adenocarcinoma epithelium. Agreement would then be the constant 1.0 or 0.8 for every patch and could rank nothing. Which property of the image it follows is a narrower question. With each patch's pixels randomly permuted, which keeps its colour and removes its morphology, the modal vote is correct for 10.8\% of the cohort and names background for 98\% of the subset patches. Agreement still clears random routing of that outcome, by $+0.030$ [0.022, 0.039], cohort-weighted. Comparing that gap with the real image's needs one scoring rule for both runs, and the real-image run kept no response text, so the comparison reads the permuted run's answers by the earlier rule too (Section~\ref{subsec:models}). By that rule the permuted run's gap is $+0.066$ and the real image adds $+0.058$ [0.034, 0.078] to it on the same patches. The $+0.030$ reads the permuted answers from their LABEL lines and the $+0.058$ is measured against the earlier rule's $+0.066$, so the two do not add up to the real-image gap of $+0.123$. Colour alone therefore carries a small part of the signal, and the spatial structure of the tissue, which the permutation removes, carries the rest. When the 360 swapped patches were sent with the image of a patch from another class, the answers matched the stored labels of the patch whose image was supplied on 98.6\% of calls, against 24.1\% for the patch they were filed under, and the consistency score correlated 0.97 with the supplied patch's stored score and $-0.10$ with its own. Agreement follows the image the model is shown.

MedGemma's stated confidence does rank under one prompt. Variant 2's own confidence, routing its own answers on the same patches in the same session, reaches a cohort AUROC of 0.612 [0.595, 0.627] on validation and 0.645 [0.629, 0.660] externally and beats random routing by $+0.112$ [0.096, 0.125] and $+0.141$ [0.126, 0.155], margins comparable to Gemma-3's. The other four variants stay between AUROC 0.488 and 0.521 and within 0.026 of random. Variant 2 is also the least accurate variant, at 27.4\% and 27.5\%, so the prompt that draws a confidence that ranks also answers worst. Rerun as variant 2 of the V5 family on a different sample, the same prompt reaches AUROC 0.620. The wording-only paraphrases do not reproduce it: across both cohorts their own confidences run from AUROC 0.397 to 0.573, and on each cohort three of the four paraphrases fall below random, by as much as $-0.107$ [$-0.118$, $-0.096$]; the fourth clears it only through its list answers, which count as errors, and falls to or below random without them (Supplementary Information). Under the V5 family as a whole, weighted by class share, the base prompt's single-query confidence clears random by $+0.049$ [0.037, 0.061] where V3's does not, and agreement and averaged confidence clear it by $+0.043$ [0.026, 0.060] and $+0.047$ [0.029, 0.064] and do not differ from each other ($-0.004$ [$-0.029$, 0.020]). Agreement is weaker there than under V3 ($+0.043$ against $+0.123$, on different samples), so the two signals meet partly because agreement falls. The uninformative stated confidence is therefore a property of MedGemma under the V3 prompt, which was chosen on macro F1 with no reference to confidence (Section~\ref{subsec:models}), and under prompts that keep V3's structure.

Treated as a probability rather than as a ranking, the consistency score is not well calibrated: its expected calibration error as a confidence value is 0.374 on the validation subset. A five-valued score cannot be a calibrated probability without rescaling, and we claim nothing about it beyond ranking, which is what a routing policy needs.

\subsection{Label-Token Against Verbalized Confidence}\label{subsec:logprob-results}

A verbalized number could fail for either of two reasons: the model may not represent its own uncertainty, or it may represent it and report it badly. Reading the label-token distribution from the same responses is meant to separate the two, and Table~S2 in the Supplementary Information gives the full comparison.

Gemma-3's two sources agree and both transfer. The verbalized number reaches AUROC 0.701 on validation and 0.709 on the external cohort, the label-token distribution 0.667 and 0.690, and both route above random on both cohorts by $+0.106$ to $+0.133$. The label-token signal still clears random on the patches where all nine classes reached the top 20 alternatives, by $+0.046$ [0.006, 0.087] and $+0.060$ [0.022, 0.102], so for this model the spoken number reads out a distribution that carries information.

MedGemma's distribution does not pass the same checks. On the full validation sample its label-token signal beats the verbalized number by $+0.057$ [0.030, 0.082] and clears random by $+0.049$ [0.028, 0.069], where the verbalized number does not clear random ($-0.008$ [$-0.021$, 0.004]). That advantage comes from patches whose top 20 alternatives omitted a class: it is $+0.084$ [0.046, 0.123] over the verbalized number where eight classes appeared, while on the 937 patches where all nine appeared the label-token signal falls below the verbalized number by $-0.076$ [$-0.108$, $-0.044$] and below random by $-0.064$ [$-0.089$, $-0.040$]. Fewer classes in the top 20 go with both a higher renormalized confidence and a higher accuracy (45.1\% correct where seven classes appeared, 39.5\% with eight, 36.9\% with nine), so part of the validation advantage is the class count rather than the probability the model assigned to its answer, and it cannot be read as ranking information that the stated number discards. On the external cohort the label-token signal falls below the verbalized one on the full sample, by $-0.077$ [$-0.103$, $-0.053$], and below random, by $-0.056$ [$-0.076$, $-0.037$], and both reversals are larger where all nine classes appeared ($-0.112$ [$-0.145$, $-0.080$] and $-0.099$ [$-0.122$, $-0.072$]).

Neither half of MedGemma's label-token result is secure. The validation advantage does not survive the censoring check, and the external reversal, which does, comes from 25 slides: under the design effect of Section~\ref{subsec:routing-results} the gap below random stops excluding zero at an intra-cluster correlation of 0.09 and the reversal against the verbalized number at 0.11 (Table~S1). We therefore read the external reversal as a point estimate.

The mechanism behind the external inversion is a pooling effect rather than an absence of signal. Within each predicted class the label-token confidence does discriminate, reaching AUROC 0.904 within background and 0.880 within adipose on the external split. What inverts the pooled number is the ordering \emph{between} classes: MedGemma is most confident at the token level on the class it is worst at, with smooth muscle taking 57.5\% of its external-cohort predictions at mean confidence 0.942 and 13.0\% accuracy while background and adipose sit at 0.805 and 0.840 confidence with 88\% accuracy. Ranking every patch together therefore puts most of the errors at the top. The verbalized number cannot show this inversion, because a signal with 96\% of its mass on one value ranks nothing in either direction (Section~\ref{subsec:disc-routing}).

Two cautions apply to Table~S2. Its mutual information, the plug-in estimator on the 15 bins used elsewhere, is not comparable between the two sources, because the near-continuous label-token signal piles into a single bin; the comparison therefore rests on the rank statistics. And every ECE lies between 0.46 and 0.64, so neither channel could be used as a probability without rescaling.

\section{Discussion}\label{sec:discussion}

\subsection{Accuracy and Per-Class Failures}\label{subsec:disc-tuning}

MedGemma outperforms Gemma-3 by 5.2~pp on val and 4.3~pp on test. The gap narrows on the external cohort, which is consistent with part of MedGemma's advantage being specific to the source collection, though the reduction is small enough that we would not lean on it. The two models reach their accuracies through different per-class profiles: MedGemma dominates BACK and LYM, the structurally distinctive classes, while Gemma-3 does better on TUM and STR, which need tumour context. That is easy to mistake for the explanation of the routing divergence, and it is not. Section~\ref{subsec:disc-routing} shows that divergence comes from how widely each model spreads its confidence, which is close to independent of which classes each one gets right.

One of MedGemma's three near-zero classes, normal mucosa, moves with the prompt. On the 450 pilot patches it reaches F1 0.61 under V2 and 0.53 under V4B against 0.00 under V3, and the share of predictions that are smooth muscle ranges from 1\% under V4B to 51\% under V3; mucus stays at or below 0.12 and stroma at or below 0.18 under every version (Supplementary Information). Under V3 the smooth-muscle label takes 84\% of mucus patches, 90\% of normal mucosa and 42\% of stroma, at the same stated confidence of 0.95. Randomly drawn examples of each confusion (Supplementary Figure~S2) show mislabelled normal-mucosa patches with crypts in view and mislabelled stroma that resembles true smooth muscle, so the stroma errors look partly visual and the normal-mucosa errors look like a bias of the prompt.

At 31 to 41\% accuracy neither VLM is usable as a standalone classifier on this task. The supervised reference, at 91.1\% on the external cohort, puts a number on the distance, and zero-shot PLIP, which uses no task labels either, is 18.6 points above MedGemma there at 53.8\%.

\subsection{Why MedGemma Routing Fails and Gemma-3 Routing Works}\label{subsec:disc-routing}

The natural reading of an uninformative confidence is dilution: high confidence concentrated on a systematically wrong class, MUS, whose precision is 0.23. Arithmetic rules that out. Confidence concentrated on a class wrong 77\% of the time would drive the confidence-correctness AUROC near 0.33, not leave it at 0.499. MedGemma's confidence does not concentrate anywhere, because it barely varies: the spread of its mean across predicted classes is 0.016. A constant cannot be anti-informative and it cannot be thresholded. There is no high-confidence subset distinct from the rest, so the AUROC sits near 0.5 because of the structure of the signal rather than any relationship between confidence and error. That covers the four classes holding 80\% of predictions, where the model emits one or two values. It does not cover background and debris, where the confidence varies a little and the within-class AUROC is 0.299 and 0.533, one below chance and one above it, each with an interval excluding 0.5 (Section~\ref{subsec:routing-results}). So the signal is degenerate over most of the predictions and points no consistent way on the rest.

The distinction matters. Dilution would be a fixable problem: correct the class bias and the confidence becomes usable. Degeneracy says something else, that this model under this prompt does not vary its stated confidence enough to express uncertainty at all. The MUS over-prediction is real, it depresses accuracy, and it survived four prompt variants targeting the class. It is nonetheless a separate failure from the confidence failure, and the two should not be conflated.

So the difference between the medically-tuned model and its general-purpose base is not that one is well-calibrated and the other badly calibrated, since both are badly calibrated. It is that one emits a confidence \textit{scale} and the other emits what is close to a constant token. Gemma-3's confidence is weakly informative rather than empty (AUROC 0.70, 0.11 bits), spread over nine values with a per-class mean spanning 0.199, and two checks say it is a genuine correctness signal rather than a proxy for class identity: its within-class AUROC stays high, so it separates correct from incorrect \textit{inside} each predicted class, and it beats a routing baseline built only from predicted-class accuracies, which MedGemma does not. Even so, a selective-accuracy AUC of 0.47 against 0.35 random is modest, far below the supervised CNN, and nowhere near deployment-grade alone.

Response agreement under V3, the prompt whose stated confidence is uninformative, behaves differently, and it does so in two parts. Taking the modal label of the five variants does not raise accuracy at cohort level; over five wording-only paraphrases it gains 2.6 points on validation and 7.0 externally, because the paraphrases themselves differ in accuracy. Using the \textit{pattern of agreement} across those same queries as a routing signal does pay: weighted to the whole split it clears random on both cohorts and beats both confidence signals when they route the same outcome, though the external margins over confidence do not survive a plausible within-slide correlation (Section~\ref{subsec:consistency-results}). Averaged confidence over the same five answers is only weakly informative, and on validation only through variant 2. That split between what a model answers and what it says about the answer is the one Kadavath et al.\ found for verbalized confidence against internal probabilities \cite{kadavath2022know}, and which Xiong et al.\ and Tian et al.\ found to survive most prompting strategies for eliciting calibration \cite{xiong2024uncertainty,tian2023askcalibration}; what Wang et al.\ and Manakul et al.\ contribute is the method of reading the disagreement itself \cite{wang2023selfconsistency,manakul2023selfcheckgpt}.

Three results bound what that disagreement measures here. In the original family it is concentrated in one variant: without variant 2, agreement among the V3 edits carries a small signal on validation and none externally. Rewording alone produces it nonetheless, since five paraphrases that change only V3's wording recover, with both families read by the earlier rule, about three quarters of the original family's gain on validation and about two fifths externally. And it depends on the image: without one every variant gives the same answer on every repeat, and given another patch's image the answers follow that image; with the pixels permuted, which keeps colour and removes morphology, it still beats random by $+0.030$, so a small part of what it follows is colour. The answer moves with phrasing on the patches the model is least reliable on, and that movement is measurable without any access to the model's internals.

The per-variant breakdown bounds the claim. The variant appending an explicit instruction to lower confidence under ambiguity produced a mean confidence of 0.9511, against 0.9518 for the unmodified prompt. Asking the model more directly to express uncertainty changed nothing. Within V3, then, an explicit instruction to express uncertainty leaves the stated confidence unchanged, although a restructured prompt, variant 2, yields a stated confidence that ranks (Section~\ref{subsec:consistency-results}). The label-token distribution behind those words does not rescue the signal. Its validation advantage over the stated number reverses on the patches whose top 20 alternatives included every class, and on the external cohort it falls below random, though that result does not survive a plausible within-slide correlation (Section~\ref{subsec:logprob-results}).

\section{Limitations}\label{sec:limitations}

\textbf{The negative result is bounded by one prompt and two elicitations.} MedGemma's verbalized confidence fails under V3, the three edits of it and four wording-only paraphrases of it, and ranks under variant 2, a restructured prompt run on the same patches, and more weakly under the V5 family (Section~\ref{subsec:consistency-results}), so what we document is a property of a model and a prompt together. That is enough to show the failure is not fixed in the model and not enough to say which prompt is typical; we never swept the prompt space with confidence quality as the criterion, and a study that did could place V3 anywhere from common to unusual. The same bound applies to the elicitation. We measured the number the model writes and the label-token distribution behind it, which is two channels out of many: not sampled self-consistency at non-zero temperature, not sequence-level probabilities, not a confidence asked for in a separate call, and not a trained probe on hidden states, which the endpoint does not expose. Within the K=5 set the bound is tighter still. Four of the five variants are edits of V3 and only variant 2 descends from V4, and within that family the agreement signal sits almost entirely in whether variant 2 agrees with the others (Table~\ref{tab:t6d}). The wording-only family recovers most of the signal on validation without that restructuring, but externally it recovers about two fifths of it, and that external gap does not survive a plausible within-slide correlation. Those five confidences are also not commensurable: variant 4 instructs the model to lower its confidence under ambiguity, so averaging across variants pools values elicited on different scales. Here the pooling changes little, since variant 4's mean confidence is 0.9511 against 0.9518 for the unmodified prompt, but the mean-of-5 signal is a pooled quantity rather than the model's confidence. Read the result as an existence proof. A prompt selected on macro F1 can produce a confidence signal worthless for ranking, and macro F1 gives no warning of it.

\textbf{Intervals are unclustered, and the substitute for clustering is not equivalent to it.} Every interval here treats patches as independent when validation patches nest within 86 slides and external-test patches within 25. We cannot do better: MedMNIST v2 publishes no map from a PathMNIST array index to its source slide, and the distributed archives hold only images and labels. Table~S1 shows which conclusions survive. MedGemma's validation routing gap does not exclude zero even unclustered, and its external gap breaks at $\rho = 0.09$; the external margins of agreement over the two confidence signals break at $\rho = 0.07$ and 0.19 and the external label-token results at 0.09 and 0.11, and patches cut from one of 25 slides plausibly correlate that much. The design-effect model also assumes equal cluster sizes at the observed mean, a single correlation shared across classes and models, and widening that leaves each interval centred where it was. Real slides are unequal and their correlation almost certainly varies by tissue class, so every break-even value is an order of magnitude rather than a threshold. Multiplicity is handled separately, by a Holm adjustment over the contrasts the conclusions rest on (Section~\ref{subsec:routing-methods}): every one of the 19 holds except MedGemma's validation routing gap and agreement among the three V3 edits on the external split, both null before adjustment as well. Holding under the adjustment does not make a contrast a claim: MedGemma's two validation label-token contrasts hold and are not claimed, for the censoring reason above. The adjustment treats each contrast's interval as given, so it inherits the clustering problem rather than addressing it.

\textbf{Tie-dominated signals limit what any ranking metric can say.} Verbalized confidence takes between two and ten distinct values depending on model and prompt, and 92\% of MedGemma's patch pairs are tied. On such a signal AUROC is pinned near 0.5 and every AUC depends on how ties are broken. We resolve them by expectation and report the induced spread, but the deeper point survives the convention: this signal does not have the resolution to support fine-grained ranking claims, and neither does a five-valued consistency score treated as a probability rather than an order.

\textbf{Two analyses rest on samples rather than the full splits.} The K=5 experiment ran on 1,800 patches per split, drawn to balance correct and incorrect predictions within each class, and its cohort estimates come from inverse-probability weights. The weights are exact, but the estimates inherit their variability, which the cell-wise bootstrap carries into every interval, and the strata were defined on the stored full-scale run rather than on the in-session single query that serves as the baseline. A richer agreement signal, embedding-space dispersion across the K rationales, is closed to us because the K=5 runs stored labels and confidences but not the rationale text. Table~S2 rests on 2,001 validation and 2,000 external patches, because each patch costs a billed call. The two confidence sources are paired on identical responses, so the contrast between them holds on those patches; what would move on a different draw is the pooled external number, since it turns on the realized class mix. The label-token confidence is further conditioned on which classes the endpoint returned among its top 20 alternatives, and the probability mass outside them cannot be recovered from the stored outputs (Section~\ref{subsec:logprob-methods}).

\textbf{Three groups of runs keep labels from an earlier parser.} The original-family K=5 runs, the V5 family and the label-token runs stored no response text, so their labels come from the whole-text search that misread 1.9\% of MedGemma's full-scale validation labels and 1.7\% of its external labels (Section~\ref{subsec:models}). The direction of the error on agreement can be estimated from the image-swap control, which sent the same five prompts with 360 real images and kept its text: reading its answers from the LABEL line rather than by the earlier rule raises agreement's gap over random routing, scored against the class of the image supplied, from $+0.108$ to $+0.153$, a difference of $+0.045$ [0.019, 0.074] (unweighted, 1,000 paired resamples). If the stored runs behave alike, the agreement results of Tables~\ref{tab:t6b} to~\ref{tab:t6d} understate the signal rather than inflate it. The misread answers in that control all come from the four V3-style variants, between 10 and 38 of 360 each. Both rules read every one of variant 2's answers alike, so its own confidence scores the same under either (AUROC 0.586, gap over random $+0.091$ [0.052, 0.127]), and without variant 2 agreement clears random by $+0.049$ on the LABEL line against $+0.030$ by the earlier rule (Supplementary Information). On these 360 patches, then, neither variant 2's own confidence (Table~S9) nor its share of agreement depends on the parser, though at this size the comparison is indicative only. The label-token comparison is exposed twice over: its confidence is the probability of the predicted class (Section~\ref{subsec:logprob-methods}), so a misread label changes which class's probability is taken as well as which answers count as correct.

\textbf{The supervised reference is usable only on the external cohort.} PathMNIST partitions at the patch level and publishes no slide identifiers, so the CNN's validation figures are leakage-inflated and no slide-disjoint split can be built from the distributed data. The CNN's cross-cohort drop is therefore unknown, and the paper makes no claim about whether the VLMs or the CNN are more robust to the shift. Its external accuracy, calibration and routing are clean, but they come from one training run at seed 42 and carry no variance estimate.

\textbf{Neither the endpoint nor the setting is reproducible in the strict sense.} Inference ran against a remote OpenAI-compatible endpoint that exposes no model snapshot or version identifier. Two runs of the identical prompt at temperature 0.0 disagreed on 3.0\% of validation patches and 0.0\% of external-test patches, which no single stable configuration produces, and a third session in October disagreed with the stored run on 0.9\% of external-test patches. The provider is named in Section~\ref{subsec:consistency-methods}, request-level timestamps exist only for the log-probability, V3W and image-control runs, and the served model version cannot be pinned by us or by anyone repeating this; the endpoint has since stopped serving Gemma-3-27b-it altogether. The same gap bounds the model comparison. MedGemma and Gemma-3 differ in medical tuning and in whatever else separates the two served checkpoints, so the dissociation between their confidence signals is consistent with an effect of tuning but does not establish one. Attributing it would take identified checkpoints run under one controlled configuration. The clinical setting is simulated throughout: no pathologist reviewed the routed patches, the specialist queue is exclusion from the auto-confirmed set, and whether routing reduces diagnostic error in practice is untested. All results are patch-level, colorectal, H\&E, and confined to the nine-class NCT-CRC taxonomy, with no whole-slide aggregation and no evidence about other organs, stains or label vocabularies.

\section{Conclusion}\label{sec:conclusion}

We asked whether an open medical VLM's verbalized confidence carries enough information to drive selective prediction on colorectal H\&E patches, evaluating MedGemma-27b-it against its base Gemma-3-27b-it with a supervised ResNet-18 as reference, and CRC-VAL-HE-7K as an external cohort differing in collection period, cohort and scanning batch. Both collections were Macenko colour-normalized before release, so the shift is what survives that normalization rather than a raw difference in staining. The framing is diagnostic: at 31--41\% zero-shot accuracy neither VLM is a usable classifier, so the question is not whether to deploy one but what its stated confidence is worth.

The answer differs sharply by model, and the reason is more basic than calibration. MedGemma's verbalized confidence takes four distinct values on validation and five on the external cohort, with 96\% of patches at one of them, which leaves the confidence-correctness AUROC near 0.5 (0.499 validation, 0.528 external test), almost no ranking information whatever its interval, and the mutual information at the estimator's floor. Routing by it moves the selective-accuracy AUC by $-0.004$ on validation, indistinguishable from random, and $+0.037$ on the external cohort, and it is beaten on both by a baseline that uses only the predicted class and ignores the confidence entirely. Gemma-3 spreads its confidence across nine values, reaches AUROC 0.70, and routes reliably if weakly on both cohorts (gaps $+0.114$ and $+0.115$). That a medically tuned model's confidence is less usable than its base's under the same prompt and endpoint is the first finding, though without pinned checkpoints it cannot be credited to tuning alone, and its mechanism is degeneracy rather than misdirection: MedGemma does not concentrate confidence on a wrong class, it emits what is nearly a constant, and an instruction to lower that confidence under ambiguity moved its mean by 0.0007.

Reading the label-token distribution from the same responses, Gemma-3's spoken number reads out a distribution that ranks on both cohorts. MedGemma's distribution does not rescue its number. Its apparent validation advantage comes from patches on which fewer than nine classes reached the top 20 alternatives and reverses on the patches whose top 20 alternatives happened to include all nine, and on the external cohort the token signal falls below random, because the model is most confident at the token level on the class it is worst at; that reversal does not survive a within-slide correlation of about 0.1.

Response agreement across five prompt variants routes where the stated number does not. Weighted to the whole split, it clears random by $+0.123$ on validation and $+0.106$ externally and beats both confidence signals routing the same outcome, though the external margins over confidence do not survive the clustering correction (Section~\ref{subsec:consistency-results}). In that family most of the signal comes from one restructured variant, but five paraphrases that change only V3's wording recover about three quarters of it on validation and about two fifths externally, and the signal requires the image, though a small part of it survives when only the image's colour is kept. Majority voting over the original variants does not raise accuracy, so there the value lies in the disagreement pattern and not in the extra votes. Both results are scoped to the V3 prompt, chosen on macro F1: on the same patches a restructured variant's stated confidence routes about as well as Gemma-3's, though its AUROC is lower and that variant is the least accurate of the five.

The practical lesson is narrow. A prompt selected for accuracy can leave a model's stated confidence with almost nothing to rank, macro F1 gives no warning of it, and an explicit instruction to express uncertainty does not repair it. The confidence signal has to be measured directly, on the prompt that will be used, before anything is built on it.

\section*{Declarations}

\begin{itemize}
\item \textbf{Funding:} No external funding was received for this study.
\item \textbf{Conflict of interest/Competing interests:} The authors declare no competing interests.
\item \textbf{Ethics approval and consent to participate:} Not applicable. This study uses only the publicly available, de-identified PathMNIST benchmark derived from NCT-CRC-HE-100K and CRC-VAL-HE-7K; no new human subjects data were collected.
\item \textbf{Consent for publication:} Not applicable.
\item \textbf{Supplementary information:} The Supplementary Information reports in full the design-effect sensitivity that stands in for clustered intervals (Table~S1), the label-token against verbalized confidence comparison (Table~S2), the prompt-selection pilot and the validation results without it (Tables~S3 and S4), the binary negative log-likelihood against temperature (Figure~S1), the label-token comparison by number of classes returned (Table~S5), the multiplicity adjustment (Table~S6), the texts of the wording-only prompt family and of the original variants with the parse outcome of every variant, the PLIP and QuiltNet baselines (Table~S7), calibration maps beyond temperature scaling (Table~S8), each prompt variant's own stated confidence (Table~S9), agreement against the number of prompts (Table~S10), operating points and query cost (Table~S11), per-class results under each pilot prompt with example misclassified patches (Table~S12, Figure~S2), and the inference configuration (Table~S13). Their findings are stated in the main text at the points where they bear on a claim.
\item \textbf{Data availability:} PathMNIST is publicly available as part of MedMNIST v2 \cite{yang2023medmnist}, and its 64 and 224 px images as part of MedMNIST+ \cite{doerrich2025medmnistplus}. Both are derived from NCT-CRC-HE-100K and CRC-VAL-HE-7K \cite{kather2018nct,kather2019predicting}, released under CC BY 4.0.
\item \textbf{Materials availability:} Not applicable.
\item \textbf{Code availability:} The analysis code, the evaluation scripts and the saved model predictions that every table and figure is computed from are archived on Zenodo as version v7.0, \href{https://doi.org/10.5281/zenodo.24354813}{https://doi.org/10.5281/zenodo.24354813}, the release described here; the concept DOI \href{https://doi.org/10.5281/zenodo.22245989}{10.5281/zenodo.22245989} resolves to the most recent version. The archive holds the saved outputs that every table and figure is computed from, so the analysis scripts run without API calls, and the README maps each script to the table or figure it produces. Development history is at \href{https://github.com/anayy09/cpath-triage}{https://github.com/anayy09/cpath-triage}.
\item \textbf{Author contributions:} A.S. (Anay Sinhal) conceived the study,
designed the experiments, implemented the evaluation pipeline and all models,
conducted the analysis, and wrote the manuscript. Am.S. (Amit Sinhal), A.P.
(Ashish Pandey), and Ar.S. (Arpana Sinhal) provided supervision, methodological 
guidance, and critical review of the manuscript. All authors read and approved 
the final version.
\end{itemize}

\bibliography{bibliography}

\end{document}